\documentclass[a4paper]{article}
\usepackage{graphicx}
\usepackage[utf8]{inputenc} 

\usepackage{color}
\usepackage{amssymb} 
\usepackage{amsmath}
\usepackage{hyperref}
\usepackage[capitalize]{cleveref}
\usepackage{mathtools}
\usepackage{authblk}
\usepackage[misc]{ifsym}

\newcommand{\X}{{\cal X}}
\newcommand{\Z}{{\cal Z}}

\def\R{\mathbb{R}}

\def\E{\mathbb E}

\def\vus{{\rm VUS}}

\def\I{{\mathbb I}}

\newcommand{\roc}{\rm  ROC}

\DeclarePairedDelimiter\abs{\lvert}{\rvert}%
\DeclarePairedDelimiter\norm{\lVert}{\rVert}%

\makeatletter
\let\oldabs\abs
\def\abs{\@ifstar{\oldabs}{\oldabs*}}
\let\oldnorm\norm
\def\norm{\@ifstar{\oldnorm}{\oldnorm*}}
\makeatother

\newcommand{\Prob}[2][]{ \mathbb{P}_{#1}\left( #2 \right)}

\newtheorem{theorem}{Theorem}
\newtheorem{definition}{Definition}

\newtheorem{remark}{Remark}
\newtheorem{example}{Example}

\begin{document}


%
\title{Trade-offs in Large-Scale Distributed Tuplewise Estimation and
Learning}
%
\author[1,2]{Robin Vogel (\Letter)}
\author[3]{Aur\'elien Bellet}
\author[1]{Stephan Cl\'emen\c{c}on}
\author[1]{Ons Jelassi}
\author[1]{Guillaume Papa}
 
\affil[1]{Telecom ParisTech, LTCI, Universit\'e Paris Saclay, France\\
\href{mailto:first.last@telecom-paristech.fr}{first.last@telecom-paristech.fr}}
\affil[2]{IDEMIA, France\\ \href{mailto:first.last@idemia.fr}{first.last@idemia.fr}}
\affil[3]{INRIA, France\\  \href{mailto:first.last@inria.fr}{first.last@inria.fr}}
\date{}
%
%
\maketitle              
\begin{abstract}
The development of cluster computing frameworks has
allowed practitioners to scale out various statistical estimation and machine
learning algorithms with minimal programming effort. This is especially true for machine learning problems whose objective function is nicely separable across individual data points, such as classification and regression.
In contrast, statistical learning tasks involving pairs (or
more generally tuples) of data points --- such as metric learning,
clustering or ranking --- do not lend themselves as easily to
data-parallelism
and in-memory computing.
In this paper, we investigate how to balance between statistical performance
and computational efficiency in such distributed tuplewise statistical
problems. We first propose a simple strategy based on occasionally
repartitioning
data across workers between parallel computation stages, where the number of
repartitioning steps rules the trade-off between accuracy and runtime. We then
present some theoretical results highlighting the benefits brought by the
proposed method in terms of variance reduction, and extend our results to
design
distributed stochastic gradient descent algorithms for tuplewise
empirical
risk minimization.
Our results are supported by numerical experiments in pairwise
statistical estimation and learning on synthetic and real-world datasets.

\end{abstract}

%
%
%


\section{Introduction}

Statistical machine learning has seen dramatic development over the last
decades.
The availability of massive datasets combined with the increasing need to
perform predictive/inference/optimization tasks in a wide variety of domains
has given a considerable boost to the field and led to successful applications. 
In parallel, there has been an ongoing technological progress in the
architecture of data repositories and distributed systems, allowing to
process ever larger (and possibly complex, high-dimensional) data sets
gathered on distributed storage platforms. 
This trend is illustrated by the development of many
easy-to-use cluster computing frameworks for large-scale distributed data
processing. These frameworks implement the data-parallel setting, in which
data points are partitioned across different machines which operate on their
partition in parallel. Some striking examples are Apache
Spark
\cite{Zaharia2012c} and Petuum \cite{petuum}, the latter being
fully targeted to machine learning.
The goal of such frameworks is to abstract away the network and
communication aspects in order to ease the deployment of distributed
algorithms on large computing clusters and on the cloud, at the cost of some
restrictions in the types of operations and parallelism that can be
efficiently
achieved.
However, these limitations as well as those arising from network latencies or the nature of certain memory-intensive operations are often ignored or incorporated in a stylized manner in the mathematical description and analysis of statistical learning algorithms (see \textit{e.g.}, \cite{Balcan2012a,Daume-III2012a,Bellet2015b,arjevani2015communication}).
The implementation of statistical methods proved to be theoretically sound may
thus be hardly feasible in a practical distributed system, and seemingly minor
adjustments to scale-up these procedures can turn out to be disastrous in
terms of statistical performance, see e.g. the discussion in \cite{Jordan13}.
This greatly restricts their practical interest in some applications and urges
the statistics and machine learning communities to get involved with distributed computation more deeply \cite{BBL11}.

 In this paper, we propose to study these issues in the context of 
 \textit{tuplewise} estimation and learning problems, where the statistical
 quantities of interest are not basic sample means but come in the
 form of averages over all pairs (or more generally, $d$-tuples) of data
 points. Such data functionals are known as
 $U$-statistics \cite{Lee90,PenaGine99}, 
 and many empirical quantities describing global properties of a probability
 distribution fall in this category (\textit{e.g.}, the sample
 variance,
 the Gini mean difference, Kendall's tau coefficient). $U$-statistics are
 also
 natural empirical
 risk measures in several learning problems such as ranking \cite{CLV08},
 metric
 learning \cite{Vogel2018a}, cluster analysis 
 \cite{CLEM14} and risk assessment \cite{bertail2006incomplete}.
 The behavior of these statistics is well-understood and a sound theory for
 empirical risk minimization based on $U$-statistics is now documented in the
 machine learning literature \cite{CLV08}, but the computation of a
 $U$-statistic poses a serious scalability challenge as it involves a
 summation over an exploding number of pairs (or $d$-tuples) as the dataset
 grows in size.
 In the centralized (single machine) setting, this can be
 addressed by appropriate subsampling methods, which have been shown to achieve a nearly
 optimal balance between computational cost and statistical accuracy 
 \cite{CBC2016}. Unfortunately, naive implementations in the case of a
 massive distributed dataset either greatly damage the accuracy or are
 inefficient due to a lot of network communication (or disk I/O).
 This is due to the fact that, unlike basic sample means,
 a $U$-statistic is not separable across the data partitions.

 Our main contribution is to design and analyze distributed methods
 for statistical estimation and learning with
 $U$-statistics that
 guarantee a good trade-off between accuracy and scalability. Our approach
 incorporates an occasional data repartitioning step between parallel
 computing stages in order to circumvent the limitations induced by data
 partitioning over the cluster nodes. The number of repartitioning steps
 allows to trade-off between statistical accuracy and
 computational efficiency.
 To shed light on this phenomenon, we first study the setting of
 statistical estimation, precisely quantifying the variance of estimates
 corresponding to several strategies. Thanks to the use of Hoeffding's decomposition 
 \cite{Hoeffding48}, our
 analysis reveals the role
 played by each component of the variance in the effect of repartitioning. We
 then discuss the extension of these results to statistical
 learning and design efficient and scalable stochastic gradient
 descent algorithms for distributed empirical risk minimization.
 Finally, we  carry out some numerical experiments on pairwise estimation and
 learning tasks on synthetic and real-world datasets to support our results from an empirical perspective.

The paper is structured as follows. Section~\ref{sec:background}
reviews background on $U$-statistics and their use
in statistical estimation and learning, and discuss the common practices in
distributed data processing. Section~\ref{sec:estimation} deals with
statistical tuplewise estimation: we introduce our general approach for the
distributed setting and derive (non-)asymptotic results describing its
accuracy. Section~\ref{sec:erm} extends our approach to statistical tuplewise
learning. We provide experiments supporting our results in Section~\ref{sec:experiments},
and we conclude in Section~\ref{sec:conclusion}.
Proofs, technical details and additional results can be found in the
supplementary material.


\section{Background}\label{sec:background}

In this section, we first review the definition and properties of
$U$-statistics, and discuss some popular applications in statistical
estimation and learning. We then discuss the recent randomized methods
designed
to scale up tuplewise statistical inference to large datasets stored on a
single machine. Finally, we describe the main features of cluster
computing frameworks.

\subsection{$U$-Statistics: Definition and Applications}
\label{sec:ustat-intro}

$U$-statistics are the natural generalization of i.i.d. sample means 
to tuples of points. We state the definition of $U$-statistics in their
generalized form, where points can come from $K\geq 1$ independent samples.
Note that we recover classic sample mean statistics in the case where
$K=d_1=1$.

\begin{definition}\label{def:Ustat}{\sc (Generalized $U$-statistic)}
Let $K\geq 1$ and $(d_1,\; \ldots,\; d_K)\in \mathbb{N}^{*K}$. For each $k\in
\{1,\dots,K\}$, let $
\mathbf{X}_{\{1,\;\ldots,\; n_k  \}}=(X^{(k)}_{1},\;\ldots,\; X^{(k)}_{n_k})$ be an independent sample of size
$n_k\geq d_k$ composed of i.i.d. random variables with values
in some measurable space $\X_k$ with distribution $F_k(dx)$.
Let  $h:\X_1^{d_1}\times \cdots \times \X_K^{d_K}\rightarrow\mathbb{R}$ be a
measurable function, square integrable with respect to the probability
distribution $\mu=F_1^{\otimes d_1}\otimes \cdots \otimes F_K^{\otimes d_K}$.
Assume w.l.o.g. that $h(\mathbf{x}^{(1)},\;
\ldots, \; \mathbf{x}^{(K)})$ is symmetric within each block of arguments $\mathbf{x}^{(k)}$ (valued in $\X^{d_k}_k$).
The generalized (or $K$-sample) $U$-statistic of degrees $(d_1,\; \ldots,\; d_K)$ with kernel $H$ is defined as
\begin{equation}\label{eq:UstatG}
U_{\mathbf{n}}(h)=\frac{1}{\prod_{k=1}^K \binom{n_k}{d_k}}\sum_
{I_1}\ldots\sum_{I_K} h(\mathbf{X}^{(1)}_{I_1},\; \mathbf{X}^{(2)}_{I_2},\;
\ldots,\; \mathbf{X}^{(K)}_{I_K}),
\end{equation}
 where $\sum_{I_k}$ denotes the sum over all $\binom{n_k}
 {d_k}$ subsets $\mathbf{X}^{(k)}_{I_k}=( X^{(k)}_{i_1},\;\ldots,\; X^{(k)}_{i_{d_k}})$ related to a set $I_k$ of $d_k$ indexes $1\leq i_1< \ldots <i_{d_k}\leq n_k$ and $\mathbf{n}=(n_1,\; \ldots,\; n_K)$.
\end{definition}
The $U$-statistic $U_{\mathbf{n}}(h)$ is known to have
minimum
variance among all
unbiased estimators of the parameter 
$\mu(h)=\mathbb{E}\big[h(X^{(1)}_{1},\;\ldots,\; X^{(1)}_{d_1},\; \ldots,\;
X^{(K)}_{1},\;\ldots,\; X^{(K)}_{d_K})\big]$.
The price to pay for this low variance is a complex dependence structure
exhibited by the terms involved in the average \eqref{eq:UstatG}, as each data
point appears in multiple tuples.
The (non)asymptotic behavior of $U$-statistics and $U$-processes (\textit{i.e.}, collections of $U$-statistics indexed by classes of kernels) can be investigated by means of linearization techniques \cite{Hoeffding48} combined with decoupling methods \cite{PenaGine99}, reducing somehow their analysis to that of basic i.i.d. averages or empirical processes. One may refer to \cite{Lee90} for an account of the asymptotic theory of $U$-statistics, and to \cite{VVaart} (Chapter 12 therein) and \cite{PenaGine99} for nonasymptotic results.

$U$-statistics are commonly used as point estimators for inferring certain
global properties of a probability distribution as well as in statistical
hypothesis testing. Popular examples include the (debiased) {\it sample
variance}, obtained by setting $K=1$, $d_1=2$ and $h(x_1,x_2)=(x_1-x_2)^2$,
the {\it Gini mean difference}, where $K=1$, $d_1=2$ and $h(x_1,x_2)=\vert
x_1-x_2\vert$, and {\it Kendall's tau rank correlation}, where
$K=2$, $d_1=d_2=1$ and $h((x_1, y_1),(x_2, y_1))=\mathbb{I}\{(x_1-x_2)\cdot (y_1-y_2) >0\}$.


$U$-statistics also correspond to empirical risk
measures in statistical learning problems such as clustering \cite{CLEM14},
metric learning \cite{Vogel2018a} and multipartite ranking 
\cite{ClemRob14}. The
generalization ability of minimizers of such criteria over a class $
\mathcal{H}$ of
kernels can be derived from probabilistic upper bounds for the maximal
deviation of collections of
centered $U$-statistics under appropriate complexity conditions on $
\mathcal{H}$ (\textit{e.g.}, finite {\sc VC} dimension) \cite{CLV08,CBC2016}.
Below, we describe the example of multipartite ranking
used in our numerical experiments (Section~\ref{sec:experiments}). We
refer to \cite{CBC2016} for details on more learning problems
involving $U$-statistics.

\begin{example}[Multipartite Ranking]
\label{ex:ranking}
Consider items described by a random vector of features $X\in \mathcal{X}$
with associated ordinal labels $Y\in\{1,\ldots,K\}$, where $K\geq 2$. The goal
of multipartite ranking is to learn to rank items in the same preorder as that defined by the labels, based on a training set of labeled examples.
Rankings are generally defined through a scoring function $s:\mathcal{X}\rightarrow \mathbb{R}$ transporting the natural order on the real line onto $\mathcal{X}$. Given $K$ independent samples, the empirical ranking performance of $s(x)$ is evaluated by means of the empirical $\vus$ (Volume Under the $\roc$ Surface) criterion \cite{ClemRob14}:
\begin{equation}\label{eq:emp_vus}
\widehat{VUS}(s)=\frac{1}{\prod^{K}_{k=1}n_{k}}\sum^{n_{1}}_{i_{1}=1}\ldots\sum^{n_{K}}_{i_{K}=1}\I\{s(X^{(1)}_{i_{1}})<\ldots<s(X^{(K)}_{i_{K}})\},
\end{equation}
which is a $K$-sample $U$-statistic of degree $(1,\ldots, 1)$ with kernel $h_s(x_1,\; \ldots,\; x_K)=\mathbb{I}\{s(x_{1})<\ldots<s(x_{K})   \}$.
\end{example}

\subsection{Large-Scale Tuplewise Inference with Incomplete $U$-statistics}\label{sec:incomplete_stats}

The cost related to the computation of the $U$-statistic \eqref{eq:UstatG}
rapidly explodes as
the sizes of the samples increase. Precisely, the number of terms involved in
the summation is $\binom{n_1}{d_1}\times \cdots \times  \binom{n_K}{d_K}$,
which is of order $O(n^{d_1+\ldots+d_K})$ when the $n_k$'s are all asymptotically equivalent.
Whereas computing $U$-statistics based on subsamples of smaller size would
 severely increase the variance of the estimation, the notion of 
\textit{incomplete generalized $U$%
-statistic} \cite{Blom76} enables to significantly mitigate this computational problem while maintaining a good level of accuracy.


\begin{definition}{\sc (Incomplete generalized $U$-statistic)}
\label{def:UstatI}
Let $B\geq1$. The incomplete version of the $U$-statistic \eqref{eq:UstatG} based on $B$ terms is defined by:
\begin{equation}\label{eq:UstatI}
\widetilde{U}_{B}(H)=\frac{1}{B}\sum_{I=(I_1,\;\ldots,\, I_K)\in\mathcal{D}_B}
h(\mathbf{X}^{(1)}_{I_1},\;\ldots,\; \mathbf{X}^{(K)}_{I_K})
\end{equation}
where $\mathcal{D}_B$ is a set of cardinality $B$ built by sampling uniformly with replacement in the set $\Lambda$ of vectors of tuples
$((i^{(1)}_1,\;\ldots,\; i^{(1)}_{d_1}),\; \ldots,\; (i^{(K)}_1,\;\ldots,\; i^{(K)}_{d_K}))$, where $1\leq i^{(k)}_1<\ldots<i^{(k)}_{d_k}\leq n_k$ and $1\leq k \leq K$.
\end{definition}%
Note incidentally that the subsets of indices can be selected by means of
other sampling schemes \cite{CBC2016}, but sampling
with replacement is often preferred due to its simplicity. In practice, the
parameter $B$ should be picked much smaller than the total number of tuples to reduce the computational cost.
Like \eqref{eq:UstatG}, the quantity \eqref{eq:UstatI} is an unbiased estimator of $\mu (H)$ but its variance is naturally larger:
\begin{equation}\label{eq:var_incomp}
\mathrm{Var}(\widetilde{U}_{B}(h))=\Big(1-\frac{1}{B}\Big)\mathrm{Var}(U_{
\mathbf{n}%
}(h))+
\frac{1}{B}\mathrm{Var}(h(X^{(1)}_1,\; \ldots,\; X^{(K)}_{d_K})).
\end{equation}

The recent work in \cite{CBC2016} has shown that the maximal deviations
between $\eqref{eq:UstatG}$ and \eqref{eq:UstatI} over a class of kernels $
\mathcal{H}$ of controlled complexity decrease at a rate of order $O(1/\sqrt{B})$ as  $B$ increases. An
important consequence of this result is that sampling $B=O(n)$ terms is
sufficient to preserve the learning rate of order $O_{\mathbb{P}}(
\sqrt{\log n /n})$ of the 
minimizer of the complete risk \eqref{eq:UstatG}, whose computation requires to average $O(n^{d_1+\ldots+d_K})$ terms.
In contrast, the distribution of a complete $U$-statistic built from subsamples of reduced sizes $n'_k$ drawn uniformly at random is quite different from that of an incomplete $U$-statistic based on $B = \prod_{k=1}^K \binom{n'_k}{d_k}$ terms sampled with replacement in $\Lambda$, although they involve the summation of the same number of terms. Empirical minimizers of such a complete $U$-statistic based on subsamples achieve a much slower learning rate of $O_{\mathbb{P}}(\sqrt{\log(n)/n^{1/(d_1+\ldots+d_K)} })$.
We refer to \cite{CBC2016} for details and additional results.

We have seen that approximating complete $U$-statistics by
incomplete ones is a theoretically and practically sound approach to tackle
large-scale tuplewise estimation and learning problems. However, as we shall
see later, the implementation is far from straightforward when data
is stored and processed in  standard distributed computing frameworks, whose
key features
are recalled below.

\subsection{Practices in Distributed Data Processing}

\emph{Data-parallelism}, i.e. partitioning the data across different machines
which operate in parallel, is a natural approach to store and efficiently
process massive datasets.
This strategy is especially appealing when the key stages of the
computation to be
executed
can be run in parallel on each partition of the data. As a matter
of fact, many estimation and
learning
problems can be reduced to (a sequence of) local computations on each
machine followed by a simple aggregation step. This is the case of gradient
descent-based algorithms applied to standard empirical risk
minimization problems, as the objective function is nicely separable across
individual data points. Optimization algorithms
operating in the data-parallel setting have indeed been largely investigated
in the machine learning community, see
\cite{BBL11,Boyd2011a,arjevani2015communication,cocoa} and references
therein for some recent work.

Because of the prevalence of data-parallel applications in large-scale machine
learning, data analytics and other fields, the past few years have seen a
sustained development of distributed data processing frameworks designed to
facilitate the implementation and the deployment on computing clusters.
Besides the seminal MapReduce framework \cite{dean2008}, which is not
suitable for iterative
computations on the same data, one can mention Apache Spark 
\cite{Zaharia2012c}, Apache Flink \cite{Flink} and
the machine learning-oriented Petuum \cite{petuum}.
In these frameworks, the data is typically first read
from a distributed file system (such
as HDFS, \textit{Hadoop Distributed File System}) and partitioned
across the memory of each machine in the form of an appropriate distributed data structure. The
user
can then easily specify a sequence of distributed computations to be performed
on this data structure (map, filter, reduce, etc.) through a simple API which hides
low-level distributed primitives (such as message passing between machines).
Importantly, these frameworks natively implement
fault-tolerance (allowing efficient recovery from
node failures) in a
way that is also completely transparent to the user.

While such distributed data processing frameworks come with a lot of benefits
for the user, they also restrict the type of computations that can be
performed efficiently on the data. In the rest of this paper, we investigate
these limitations in the context of tuplewise estimation and learning problems,
and propose solutions to achieve a good trade-off between accuracy and
scalability.

\section{Distributed Tuplewise Statistical Estimation}\label{sec:estimation}


In this section, we focus on the problem of tuplewise statistical
estimation in the distributed setting (an extension to statistical learning is
presented in Section~\ref{sec:erm}). We consider a set of $N\geq 1$ workers in
a complete network graph (i.e., any pair of workers can exchange messages).
For convenience, we assume the presence of a master node, which can be one of
the workers and whose role is to aggregate estimates computed by all workers.

For ease of presentation, we restrict our attention to the
case of two sample $U$-statistics of degree $(1,1)$ ($K=2$ and
$d_1=d_2=1$), see \cref{rem:high-order} in Section~\ref{sec:est_analysis} for extensions to the
general
case.
We denote by $\mathcal{D}_n=\left\lbrace
X_1,\ldots,X_n \right\rbrace$ the first sample and by $
\mathcal{Q}_m=\left\lbrace Z_1,\ldots,Z_m \right\rbrace$ the second sample (of
sizes $n$ and $m$ respectively). These samples are distributed across
the $N$ workers. For $i\in\{1,\dots,N\}$, we denote by $\mathcal{R}_i$ the subset
of data points held by worker $i$ and, unless otherwise noted, we assume for
simplicity that all subsets are of equal size $|\mathcal{R}_i| = \frac{n+m}{N}
\in \mathbb{N}$.
The notations $\mathcal{R}_i^{X}$ and $\mathcal{R}_i^{Z}$ respectively denote the subset of data points held by worker $i$ from 
$\mathcal{D}_n$ and $\mathcal{Q}_m$, with $\mathcal{R}_i^{X} \cup \mathcal{R}_i^{Z} = \mathcal{R}_i$. We denote their (possibly
random) cardinality by
$n_i =
|\mathcal{R}_i^X|$ and $m_i = |\mathcal{R}_i^Z|$.
Given a kernel $h$, the goal is to compute a good estimate of the parameter $U(h)=\mathbb{E}[h(X_1,Z_1)]$ while meeting some computational and communication constraints.

\subsection{Naive Strategies}
\label{sec:naive}

Before presenting our approach, we start by introducing two simple (but
ineffective) strategies
to compute an estimate of $U(h)$. The first one is to compute the
complete two-sample $U$-statistic associated with the full samples $
\mathcal{D}_n$ and $\mathcal{Q}_m$:
\begin{align}
\label{eq:ustat-K2}
U_{\mathbf{n}}(h)=\frac{1}{nm} \sum_{k=1}^n \sum_{l=1}^m h(X_k,Z_l),
\end{align}
with $\mathbf{n} = (n,m)$.
While $U_{\mathbf{n}}(h)$ has the lowest variance among all unbiased estimates
that can be computed from $(\mathcal{D}_n,\mathcal{Q}_m)$, computing it is a
highly undesirable solution in the distributed setting where each worker only
has access to a subset of the dataset. Indeed, ensuring
that each possible pair is seen by at least one worker would require massive
data communication over the network.
Note that a similar limitation holds for incomplete versions of 
\eqref{eq:ustat-K2} as defined in Definition~\ref{def:UstatI}.

A feasible strategy to go around this problem is for each worker to compute
the complete $U$-statistic associated with its local subsample $
\mathcal{R}_i$, and to send it to the master node who averages all
contributions. This leads to the estimate
\begin{align}
\label{eq:local_complete}
U_{\mathbf{n}, N}(h)=\frac{1}{N}\sum_{i=1}^N U_{\mathcal{R}_i}(h)\quad
\text{where }U_{\mathcal{R}_i}(h) =\frac{1}{n_i m_i} \sum_{k \in 
\mathcal{R}_i^{X}} \sum_{ l \in \mathcal{R}_i^{Z}} h(X_k,Z_l).
\end{align}
Note that if $ \min (n_i, m_i) = 0$, we simply set $U_{\mathcal{R}_i}(h) = 0$.

Alternatively, as the $\mathcal{R}_i$'s may be large, each worker can
compute an incomplete $U$-statistic $\widetilde{U}_{B,\mathcal{R}_i}(h)$ with
$B$ terms instead of $U_{\mathcal{R}_i}$, leading to the estimate
\begin{align}
\label{eq:local_incomplete}
\widetilde{U}_{\mathbf{n},N,B}(h)=
 \frac{1}{N}\sum_{i=1}^N \widetilde{U}_{B,\mathcal{R}_i}(h)\quad\text{where }
 \widetilde{U}_{B,\mathcal{R}_i}(h)=\frac{1}{B}\sum_{(k,l)\in\mathcal{R}_{i,B}} h(X_k,Z_l),
\end{align}
with $\mathcal{R}_{i,B}$ a set of $B$ pairs built by sampling uniformly
with replacement from the local subsample $\mathcal{R}_i^X\times
\mathcal{R}_i^Z$.

As shown in Section~\ref{sec:est_analysis}, strategies 
\eqref{eq:local_complete} and \eqref{eq:local_incomplete} have the undesirable
property that their accuracy decreases as the number of workers $N$ increases.
This motivates our proposed approach, introduced in the following section.

\subsection{Proposed Approach}

The naive strategies presented above are either accurate but very expensive 
(requiring a lot of communication across the network), or scalable but
potentially inaccurate. The approach we promote here is of disarming
simplicity and aims at finding a sweet spot between these two extremes. The
idea is based on \textit{repartitioning} the dataset a few times across
workers (we keep the repartitioning scheme abstract for now and postpone
the discussion of concrete choices to subsequent sections). By alternating
between parallel computation and repartitioning steps, one
considers several estimates based on the same data points. This allows to
observe a greater diversity of pairs and thereby refine the quality of our
final estimate, at the cost of some additional communication.

\begin{figure}[t]
 \centering
 \includegraphics[width=.9\textwidth]
 {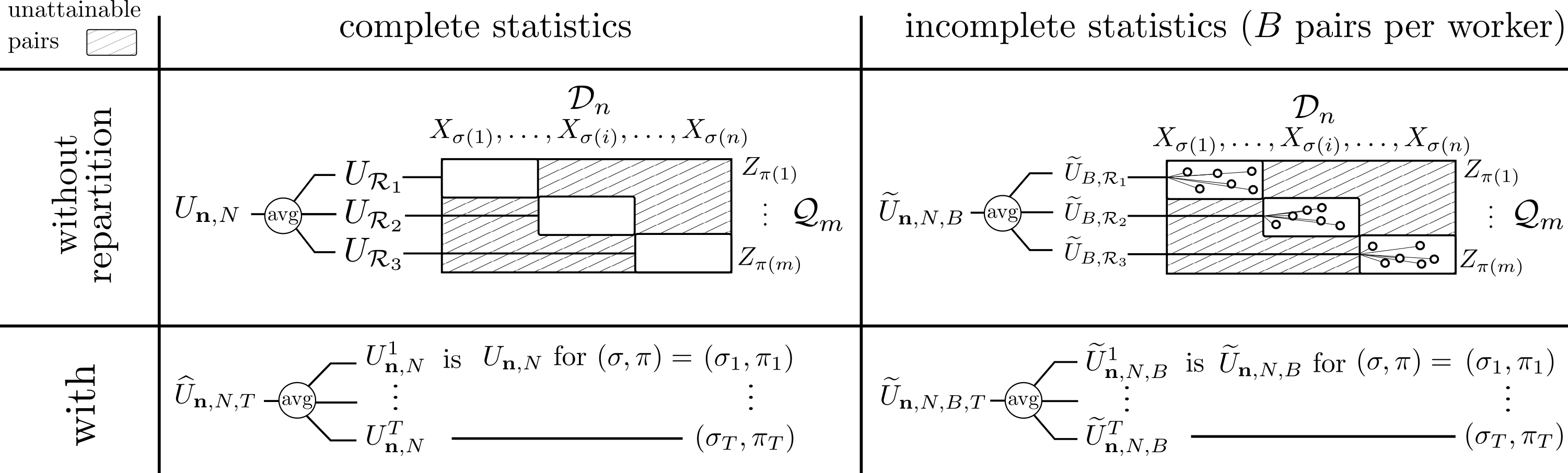}
 \caption{Graphical summary of the statistics that we compare: with/without
 repartition and with/without subsampling. Note that $\{(\sigma_t, \pi_t)\}_
 {t=1}^T$ denotes a set of $T$ independent couples of random permutations in $
 \mathfrak{S}_n \times \mathfrak{S}_m$.}
 \label{fig:est_illustr}
 \end{figure}

Formally, let $T$ be the number of repartitioning steps. We denote by $
\mathcal{R}_i^t$ the subsample of worker $i$ after the $t$-th repartitioning step, and by $U_{\mathcal{R}_i^t}(h)$ the complete $U$-statistic associated with $\mathcal{R}_i^t$. At each step $t\in\{1, \dots, T\}$, each worker $i$ computes $U_{\mathcal{R}_i^t}(h)$ and sends it to the master node. After $T$ steps, the master node has access to the following estimate:
\begin{align}
\label{eq:repart_complete}
\widehat{U}_{\mathbf{n},N,T}(h)=\frac{1}{T}\sum_{t=1}^T U_{\mathbf{n},N}^t(h),
\end{align}
where $U_{\mathbf{n}, N}^t(h)=\frac{1}{N}\sum_{i=1}^N U_{\mathcal{R}_i^t}(h)$. Similarly as before, workers may alternatively compute incomplete $U$-statistics $\widetilde{U}_{B,\mathcal{R}_i^t}(h)$ with $B$ terms. The estimate is then:
\begin{align}
\label{eq:repart_incomplete}
\widetilde{U}_{\mathbf{n}, N, B, T}(h)=\frac{1}{T}\sum_{t=1}^T\widetilde{U}_
{\mathbf{n},N,B}^t(h),
\end{align}
where $\widetilde{U}_{\mathbf{n},N,B}^t(h)=\frac{1}{N}\sum_{i=1}^N \widetilde{U}_{B,
\mathcal{R}_i^t}(h)$.
These statistics, and 
those introduced in Section~\ref{sec:naive} which do not rely on repartition, are summarized in
Figure~\ref{fig:est_illustr}.

Of course, the repartitioning operation is rather costly in terms of runtime so $T$ should be kept to a reasonably small value. We illustrate this trade-off by the analysis presented in the next section.

\subsection{Analysis}
\label{sec:est_analysis}

In this section, we analyze the statistical properties of the various
estimators introduced above. We focus here on repartitioning by 
\emph{proportional sampling without replacement} 
(prop-SWOR). Prop-SWOR creates partitions that contain the same
proportion of elements of each sample: specifically, it ensures that at any
step $t$ and for any worker $i$, $|
\mathcal{R}^t_i| = \frac{n+m}{N}$ with $|\mathcal{R}_i^{t,X}|=\frac{n}{N}$ and
$|\mathcal{R}_i^{t,Z}|=\frac{m}{N}$. We discuss the practical implementation
of this repartitioning scheme as well as some alternative choices in Section~\ref{sec:sampling}.

All estimators are unbiased when repartitioning is done with prop-SWOR. We will
thus compare their variance. Our main technical tool is a
linearization technique for $U$-statistics known as Hoeffding's
Decomposition (see \cite{Hoeffding48,CLV08,CBC2016}).

\begin{definition}{\sc (Hoeffding's decomposition)}
Let $h_1(x)=\mathbb{E}[h(x,Z_1)]$, $h_2(z)=\mathbb{E}[h(X_1,z)]$ and $h_0
(x,z)=h(x,z)-h_1(x)-h_2(z)+U(h)$. $ U_{\mathbf{n}}(h)-U(h)$ can be written as
a sum of three orthogonal terms:
\begin{align*}
 U_{\mathbf{n}}(h)-U(h)= T_n(h) + T_m(h) + W_{\mathbf{n}}(h),
\end{align*}
where $T_n(h)=\frac{1}{n}\sum_{k=1}^n h_1(X_k)-U(h)$ and $T_m(h)=\frac{1}
{m}\sum_{l=1}^n h_2(Z_l)-U(h)$ are sums of independent r.v,
while $W_{\mathbf{n}}(h)=\frac{1}{nm} \sum_{k=1}^n \sum_{l=1}^m h_0(X_k,Z_l)$
is a degenerate $U$-statistic
(\textit{i.e.}, $\mathbb{E}[h(X_1,Z_1)\vert X_1] = U(h)$ and  
$\mathbb{E}[h(X_1,Z_1)\vert Z_1] = U(h)$). 
\end{definition}

This decomposition is very convenient as the two terms $T_n(h)$ and $T_m(h)$
are decorrelated and the analysis of $W_{\mathbf{n}}(h)$ (a degenerate
$U$-statistic) is well documented \cite{Hoeffding48,CLV08,CBC2016}.
It will allow us to decompose the variance of the
estimators of interest into single-sample components $\sigma_1^2 = \text{Var}
(h_1(X))$ and $\sigma_2^2 =\text{Var}(h_2(Z))$ on the one hand, and a pairwise
component $\sigma_0^2 = \text{Var}(h_0(X_1,Z_1))$ on the other hand. Denoting
$\sigma^2=\text{Var}(h
(X_1,Z_1))$, we have $\sigma^2=\sigma_0^2 +
\sigma_1^2 + \sigma_2^2$.

It is well-known that the variance of the
complete $U$-statistic $U_{\mathbf{n}}(h)$ can be written as
$\text{Var}(U_{\mathbf{n}}(h)) = \frac{\sigma_1^2}{n} + \frac{\sigma_2^2}{m} 
+ \frac{\sigma_0^2}{nm}$  (see
supplementary material for details).
Our first result gives the variance of the estimators which do not rely on a
repartitioning of the data with respect to the variance of $U_{\mathbf{n}}
(h)$.

\begin{theorem}
\label{prop-SWOR-naive-est}
If the data is distributed over workers using prop-SWOR, we have:
\begin{align*}
    \text{Var}(U_{\mathbf{n}, N}(h)) &= \text{Var}(U_{\mathbf{n}}(h)) 
    + (N-1)\frac{\sigma_0^2}{nm} ,\\ 
    \text{Var}(\widetilde{U}_{\mathbf{n},N,B}(h)) &= \left( 1 - \frac{1}{B} \right) \text{Var}(U_{\mathbf{n}, N}(h))  + \frac{\sigma^2}{NB}.
\end{align*}
\end{theorem}

\cref{prop-SWOR-naive-est} precisely quantifies the excess variance
due to the distributed setting if one does not use repartitioning. Two
important observations are in order. First, the variance increase is
proportional to the number of workers $N$, which clearly defeats the purpose
of distributed processing. Second, this increase only depends on
the pairwise component $\sigma_0^2$ of the variance. In other words, the
average of $U$-statistics computed independently over the local partitions
contains all the information useful to estimate the single-sample
contributions, but fails to accurately estimate the pairwise
contributions. The resulting estimates thus lead to significantly larger
variance when the choice of kernel and the data distributions imply that
$\sigma_0^2$ is large compared to $\sigma_2^1$ and/or $\sigma_1^2$.
The extreme case happens when $U_{\textbf{n}}(h)$ is a degenerate
$U$-statistic,
i.e. $\sigma_1^2 = \sigma_2^2 = 0$ and $\sigma_0^2 > 0$, which is verified for example
when $h(x,z) = x\cdot z$ and $X,Z$ are both centered random variables.

We now characterize the variance of the estimators that leverage
data repartitioning steps.

\begin{theorem}
\label{prop-SWOR-redist-est}
If the data is distributed and repartitioned between workers using prop-SWOR,
we have:
\begin{align*}
    \text{Var}(\widehat{U}_{\mathbf{n},N,T}(h)) &= \text{Var}(U_{\mathbf{n}}(h))
    + (N-1)\frac{\sigma_0^2}{nmT} , \\ 
\text{Var}(\widetilde{U}_{\mathbf{n},N,B,T}(h)) &= 
\text{Var}(\widehat{U}_{\mathbf{n},N,T}(h)) - \frac{1}{TB} \text{Var}(U_{\mathbf{n}, N}(h)) + \frac{\sigma^2}{NTB}.
\end{align*}
\end{theorem}

\cref{prop-SWOR-redist-est} shows that the value of repartitioning arises from
the fact that the term accounting for the pairwise variance in $\widehat{U}_{
\textbf{n}, N, T}(h)$ is $T$ times lower than that of $U_{\mathbf{n}, N}(h)$.
This validates the fact that repartitioning is beneficial when the
pairwise variance term is significant in front of the other
terms.
Interestingly, \cref{prop-SWOR-redist-est} also implies that for a fixed
budget of evaluated
pairs, using all pairs on
each worker is always a dominant strategy over using incomplete
approximations.
Specifically, we can show that under the constraint $NBT = nmT_0/N$, $
\text{Var}(\widehat{U}_{
\mathbf{n},N,T_0}(h))$ is always smaller than $\text{Var}(\widetilde{U}_{
\mathbf{n},N,B,T}(h))$, see supplementary material for details. Note that
computing complete $U$-statistics also require fewer repartitioning
steps to evaluate the same number of pairs (i.e., $T_0\leq T$).

\begin{figure}[t]
 \centering
 \includegraphics[width=.84\linewidth]
 {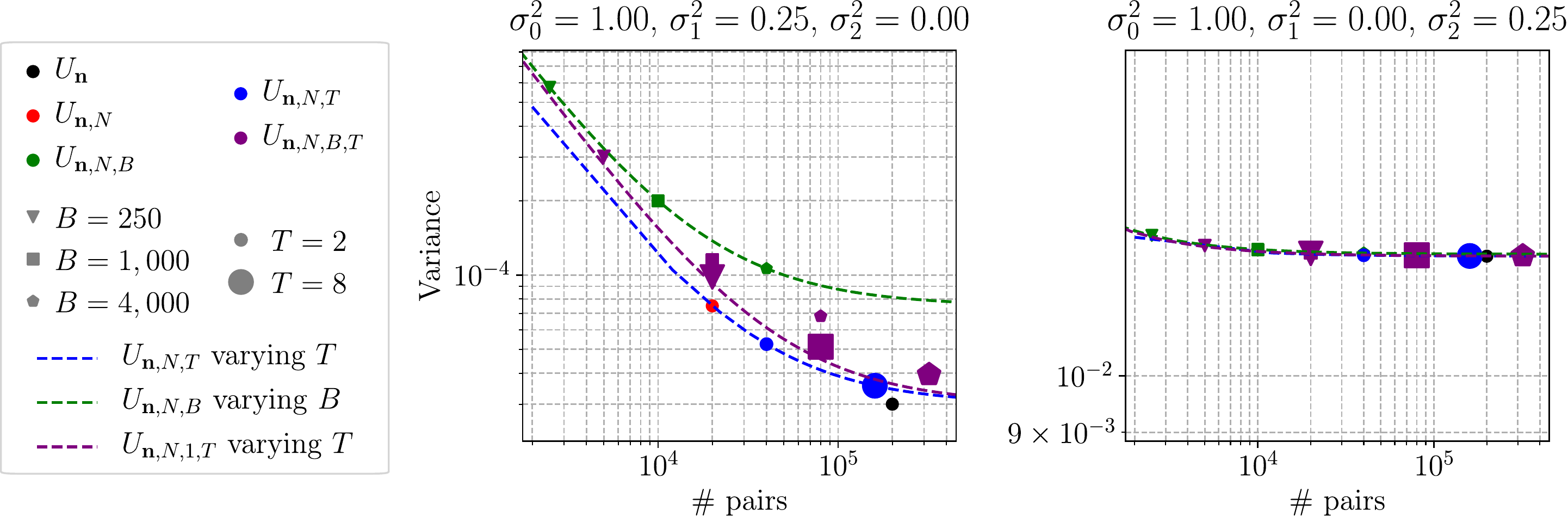}
\caption{Theoretical variance as a function of the number of evaluated pairs
for different estimators under prop-SWOR, with $n=100,000$, $m=200$
and $N=100$.}\label{fig:SWOR_theo}
\end{figure}

We conclude the analysis with a visual illustration of the variance of various
estimators with respect to the number of pairs they evaluate. We
consider the imbalanced setting where $n \gg m$, which is commonly encountered
in applications such as imbalanced classification, bipartite ranking and
anomaly detection. In this
case, it suffices that $\sigma_2^2$ be small for the influence of the pairwise
component of the variance to be significant, see \cref{fig:SWOR_theo} (left).
The figure also confirms that complete estimators dominate their incomplete
counterparts.
On the other hand, when $\sigma_2^2$ is not small, the variance of $U_{
\textbf{n}}$ mostly originates 
from the rarity of the minority sample, hence repartitioning
does not provide estimates that are significantly more accurate (see 
\cref{fig:SWOR_theo}, right).
We refer to \cref{sec:experiments} for experiments on concrete tasks with
synthetic and real data.

\begin{remark}[Extension to high-order $U$-statistics]
\label{rem:high-order}
The extension of our analysis to general $U$-statistics is
straightforward and left to the reader (see \cite{CBC2016} for a review of
the relevant technical tools). We stress the fact that the benefits of
repartitioning are even stronger for higher-order
$U$-statistics ($K>2$ and/or larger degrees) because higher-order
components of the variance are also affected.
\end{remark}

\subsection{Practical Considerations and Other Repartitioning Schemes}
\label{sec:sampling}

The analysis above assumes that repartitioning is done using prop-SWOR, which
has the advantage of exactly preserving the proportion of points from the two
samples $\mathcal{D}_n$ and $\mathcal{Q}_m$ even in the event of significant
imbalance in their size. However, a naive
implementation of prop-SWOR requires some coordination between workers at each
repartitioning step. To
avoid exchanging many messages, we propose that the workers agree
at the beginning
of the protocol on a numbering of the workers, a numbering of the points in
each sample, and a random seed to use in a pseudorandom number
generator.
This allows the workers to implement prop-SWOR without any further
coordination: at each repartitioning step, they independently draw
the same two random permutations over $\{1,\dots,n\}$ and $\{1,\dots,m\}$
using the common random seed and use these permutations to assign each point
to a single worker.

Of course, other repartitioning schemes can be used instead of prop-SWOR. A
natural choice is sampling without replacement (SWOR), which
does not require any coordination between workers. However, the partition
sizes generated by SWOR are random. This is a concern in the case of imbalanced samples, where the
probability that a worker $i$ does not get any point from the minority sample 
(and thus no pair to compute a local estimate) is non-negligible. For these
reasons, it is difficult to obtain exact and concise theoretical variances
for the SWOR case, but we show in the supplementary material that the results
with SWOR should not deviate too much from those obtained with
prop-SWOR. For completeness, in the supplementary material we also analyze
the case of
proportional sampling with replacement 
(prop-SWR): results are quantitatively similar,
aside from the fact that redistribution also corrects for
the loss of information that occurs because of sampling with replacement.

Finally, we note that deterministic repartitioning schemes may be used in
practice for simplicity. For instance, the \texttt{repartition}
method in Apache Spark relies on a deterministic shuffle which preserves the
size of the partitions.

\section{Extensions to Stochastic Gradient Descent for ERM} \label{sec:erm}

The results of Section~\ref{sec:estimation} can be extended to statistical
learning in the empirical risk minimization framework. In such problems, given
a class of kernels $\mathcal{H}$, one seeks the minimizer of 
\eqref{eq:local_complete} or \eqref{eq:repart_complete} depending on whether
repartition is used.\footnote{Alternatively, for scalability purposes, one may
instead work with their incomplete counterparts, namely 
\eqref{eq:local_incomplete} and \eqref{eq:repart_incomplete} respectively.}
Under appropriate complexity assumptions on $\mathcal{H}$ (\textit{e.g.}, of
finite {\sc VC} dimension), excess risk bounds
for such minimizers can be
obtained by combining our variance analysis of Section~\ref{sec:estimation}
with the control of maximal deviations based on Bernstein-type
concentration inequalities as done in
\cite{CLV08,CBC2016}.
Due to the lack of space, we leave the details of such analysis to the
readers and focus on the more practical scenario where the ERM problem is
solved by gradient-based optimization algorithms.

\subsection{Gradient-based Empirical Minimization of $U$-statistics}

In the setting of interest, the
class of kernels to optimize over is indexed by a real-valued parameter
$\theta\in\mathbb{R}^q$ representing the model. Adapting the notations of
\cref{sec:estimation}, the kernel $h:\X_1
\times \X_2\times\mathbb{R}^q\rightarrow\mathbb{R}$ then measures the
performance of a model $\theta\in\mathbb{R}^q$ on a given pair, and is
assumed to be convex and smooth in $\theta$. Empirical Risk Minimization (ERM) aims at finding $\theta\in\mathbb{R}^q$ minimizing 
\begin{align}
\label{eq:ustat_optim}
U_{\mathbf{n}}(\theta)=\frac{1}{nm} \sum_{k=1}^n \sum_{l=1}^m h
(X_k,Z_l; \theta).
\end{align}
The minimizer can be found by
means of Gradient Descent (GD) techniques.\footnote{When $H$ is
nonsmooth in $\theta$, a subgradient may be used instead of the gradient.}
Starting at iteration $s=1$ from an
initial model $\theta_1\in\mathbb{R}^q$ and given a learning rate $\gamma>0$,
GD consists in iterating over the following
update:
\begin{equation}
\label{eq:gd}
\theta_{s+1} = \theta_s - \gamma\nabla_{\theta} U_{\mathbf{n}}
(\theta_s).
\end{equation}
Note that the gradient $\nabla_\theta U_{\mathbf{n}}(\theta)$ is itself a
$U$-statistic with
kernel given by $\nabla_{\theta} H$, and its computation is very
expensive in the large-scale setting. In this regime, Stochastic Gradient
Descent (SGD) is a natural alternative to GD which is known to
provide a better trade-off between the amount of computation and the
performance of the resulting model \cite{Bottou2007a}.
Following the discussion of \cref{sec:incomplete_stats},
a natural idea to implement SGD is to replace
the gradient $\nabla_\theta U_{\mathbf{n}}(\theta)$ in \eqref{eq:gd} by an
unbiased
estimate
given by an incomplete $U$-statistic. The work of \cite{PCB15} shows
that SGD converges much faster than if the gradient is estimated using a
complete $U$-statistic based on subsamples with the same number of terms.

However, as in the case of estimation, the use of standard complete
or incomplete $U$-statistics turns out to be impractical in the distributed
setting.
Building upon the arguments of \cref{sec:estimation}, we propose a more
suitable strategy.

\subsection{Repartitioning for Stochastic Gradient Descent}

The approach we propose is to
alternate between SGD steps using within-partition
pairs and repartitioning the data across workers.
We introduce a parameter $n_r\in\mathbb{Z}^+$ corresponding to the number of
iterations
of SGD between each redistribution of the data. For
notational convenience, we let $r(s) := \lceil s / n_r\rceil$ so that for
any worker $i$, $\mathcal{R}^{r(s)}_i$ denotes its data partition at iteration $s\geq 1$ of SGD.

Given a local batch size $B$, at each iteration $s$ of
SGD, we propose to adapt the strategy \eqref{eq:repart_incomplete} by
having each worker $i$ compute a
local gradient estimate
using a set $\mathcal{R}^s_{i,B}$ of $B$ randomly
sampled pairs in its current local partition $\mathcal{R}^{r(s)}_i$:
$$\nabla_{\theta}\widetilde{U}_{B,\mathcal{R}^{r(s)}_i}(\theta_s)=\frac{1}{B}\sum_{
(k,l)\in\mathcal{R}^s_{i,B}} \nabla_{\theta}h(X_k,Z_l; \theta_s).$$
This local estimate is then sent to the master node who averages all contributions, leading to the following
global gradient estimate:
\begin{align}
\label{eq:sgd_estimate}
\nabla_{\theta}\widetilde{U}_{\mathbf{n},N,B}(\theta_s)=
 \frac{1}{N}\sum_{i=1}^N \nabla_{\theta}\widetilde{U}_{B,\mathcal{R}^{r(s)}_i}
 (\theta_s).
\end{align}
The master node then takes a gradient descent step as in \eqref{eq:gd} and
broadcasts the updated model $\theta_{s+1}$ to the workers.

Following our analysis in \cref{sec:estimation}, repartitioning the
data allows to reduce the variance of the gradient estimates, which is known
to greatly impact the convergence rate of SGD (see e.g. \cite{bubeck},
Theorem~6.3 therein). When $n_r = +\infty$, data is never repartitioned and
the algorithm minimizes an average of local $U$-statistics, leading to
suboptimal performance. On the other hand, $n_r=1$ corresponds to
repartitioning at each iteration of SGD, which minimizes the variance but is
very costly and makes SGD pointless. We expect the sweet spot to lie
between these two extremes: the dominance of $ \widehat{U}_{
\mathbf{n},N,T}$ over $\widetilde{U}_{\mathbf{n},N,B,T}$ established in
Section~\ref{sec:est_analysis}, combined with the common use of small
batch size $B$ in SGD, suggests that occasional redistributions are sufficient
to correct for the loss of information incurred by the partitioning of data.
We illustrate these trade-offs experimentally in the next section.




\section{Numerical Results}\label{sec:experiments}

In this section, we illustrate the importance of repartitioning for estimating
and optimizing the Area Under the ROC Curve (AUC) through a series of
numerical experiments.
The corresponding $U$-statistic is the two-sample
version of the
multipartite ranking VUS
introduced in \cref{ex:ranking} (Section~\ref{sec:ustat-intro}). The first experiment focuses on the estimation setting considered
in Section~\ref{sec:estimation}. The second experiment shows that
redistributing the data across workers, as proposed in Section~
\ref{sec:erm},
allows for more efficient mini-batch SGD.
All experiments use prop-SWOR and are conducted in a simulated
environment.

\paragraph{Estimation experiment.}

\begin{figure}[t]
    \centering
    \includegraphics[width=.77\textwidth]{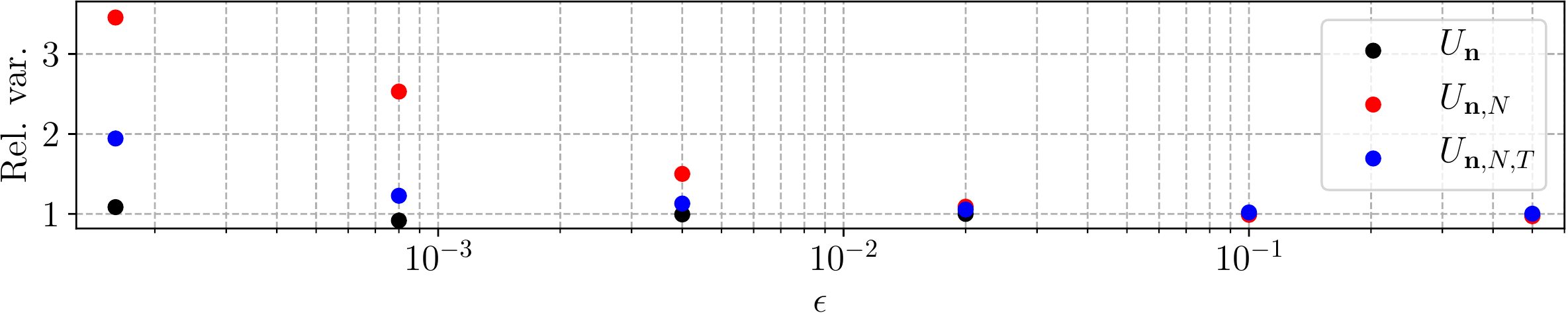}
    \caption{Relative variance estimated over $5000$ runs, $n = 5000$, $m =
    50$, $N = 10$ and $T = 4$. 
    Results are divided by the true variance of $U_{\mathbf{n}}$
    deduced from \eqref{sigmas_ext_exp} and \cref{prop-SWOR-naive-est}.}\label{fig:est_exp}
\end{figure}

\begin{figure}[t]
    \centering

    \includegraphics[width=0.65\textwidth]
    {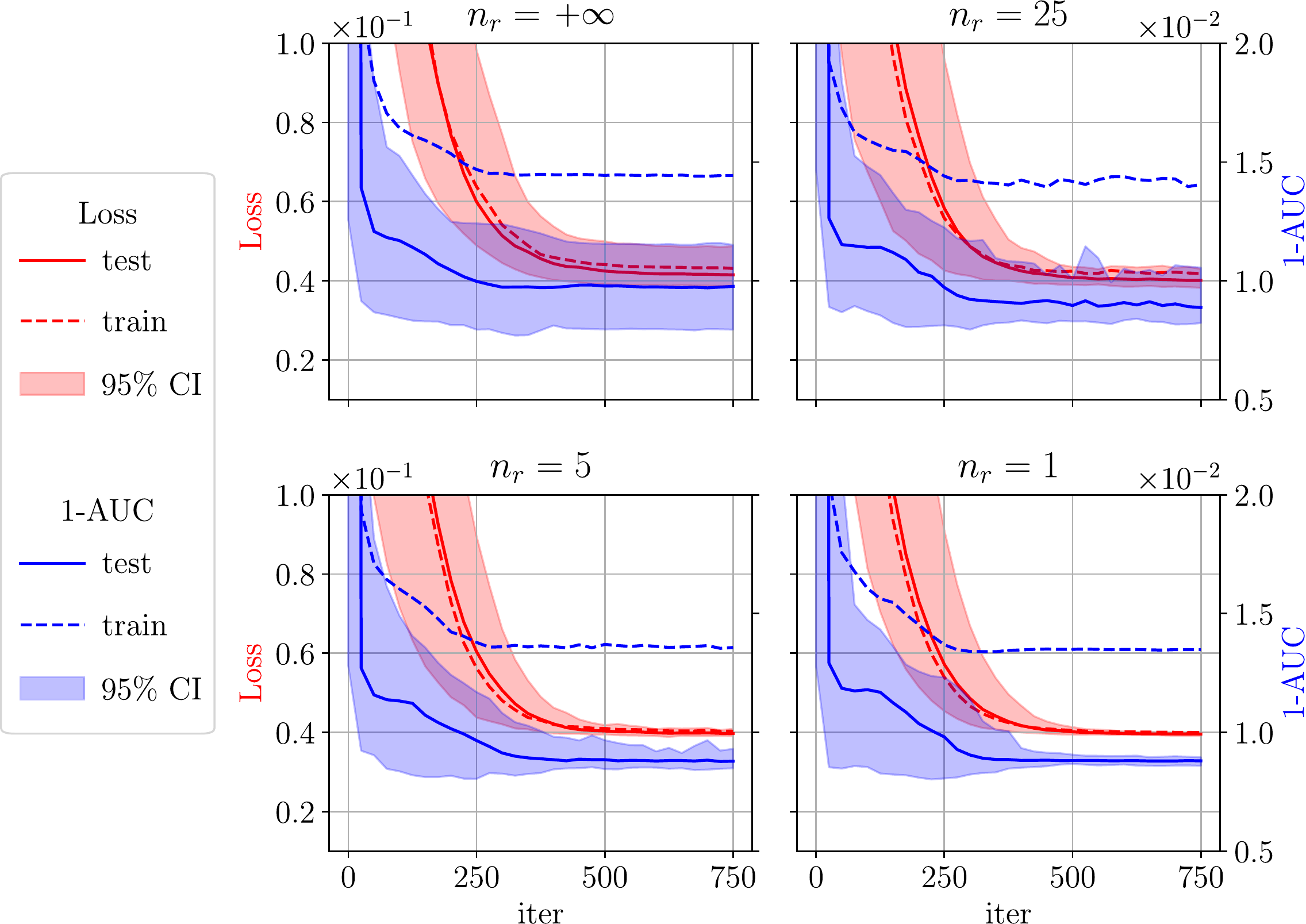}
    \caption{Learning dynamics for different repartition frequencies computed over 100 runs.}\label{fig:shuttle_optimization}
\end{figure}

We seek to illustrate the importance of redistribution for estimating
two-sample $U$-statistics with the concrete example of the AUC. The AUC
is obtained by choosing the kernel $h(x,z) = \I\{z < x\}$, and is
widely used as a performance measure in bipartite ranking and binary
classification with class imbalance.
Recall that our results of \cref{sec:est_analysis} highlighted the key role of
the pairwise component of the variance $\sigma_0^2$ being large compared to
the single-sample components.
In the case of the AUC, this happens when the data distributions are such that the expected outcome using 
single-sample information is far from the truth, e.g. in the presence of
hard pairs.
We illustrate this on simple discrete distributions for
which we can compute $\sigma_0^2$, $\sigma_1^2$ and $\sigma_2^2$ in closed
form.
Consider positive points $X\in\{0,2\}$, negative points $Z\in\{-1, +1\}$ and
$\Prob{X=2} = q$, $\Prob{Z=+1} = p$. It follows that:
\begin{align}
    \sigma_1^2 &= p^2 q (1-q),\quad \sigma_2^2 = (1-q)^2 p(1-p), 
  \text{ and } \sigma^2  = p(1-p+pq)(1-q). 
    \label{sigmas_ext_exp}
\end{align}
Assume that the scoring function has a small probability $\epsilon$
to assign a low score to a positive instance or a large score to a
negative instance. In our formal setting, this translates into letting
$p=1-q=\epsilon$ for a small $\epsilon>0$, which implies that
    $\frac{\sigma_0^2}{\sigma_1^2 + \sigma_2^2} = \frac{1-\epsilon}{2\epsilon}
        \underset{\epsilon \to 0}{\to} \infty$.
We thus expect that as the true AUC $U(h) = 1-\epsilon^2$ gets closer to $1$,
repartitioning the dataset becomes more critical to achieve good
relative precision. This is confirmed numerically, as
shown in \cref{fig:est_exp}. Note that in practice, settings
where the AUC is
very close to $1$ are very common as they correspond to well-functioning
systems, such as face recognition systems.

\paragraph{Learning experiment.}

We now turn to AUC optimization, which is the task of
learning a scoring function $s:\mathcal{X}\rightarrow\mathbb{R}$ that
optimizes the VUS criterion \eqref{eq:emp_vus} with $K=2$ in order to
discriminate between a negative and a positive class. We learn a linear
scoring function $s_{w,b}(x) = w^\top x + b$, and optimize a continuous
and convex surrogate of \eqref{eq:emp_vus} based on the hinge loss. The
resulting loss function to minimize is a two-sample U-statistic with kernel
$g_{w,b}(x,z) = \max(0, 1 + s_{w,b}(x) - s_{w,b}(z))$ indexed by the parameters $
(w,b)$ of the scoring function, to which we add a small L2 regularization term
of $0.05\norm{w}_2^2$.


We use
the shuttle dataset, a classic dataset for anomaly
detection.\footnote{\url{http://odds.cs.stonybrook.edu/shuttle-dataset/}} It contains
roughly
49,000 points in dimension 9, among which only 7\% (approx. 3,500) are
anomalies.
A high accuracy is expected for this dataset. To monitor the generalization
performance, we keep 20\% of the data as our
test set, corresponding to 700
points of the minority class and approx.
9,000 points of the majority class. The test performance is measured with
complete statistics over the 6.3 million pairs.
The training set consists of the remaining data points, which we distribute
over $N=100$ workers. This leads to approx. $10,200$ pairs per worker. The
gradient estimates are calculated following \eqref{eq:sgd_estimate} with batch
size $B=100$. We use an initial learning rate of $0.01$ with a momentum of $0.9$.
As there are more than 100 million possible pairs in the
training
dataset,
we monitor the training loss and accuracy on a fixed subset of $4.5\times
10^5$ randomly sampled pairs.


\cref{fig:shuttle_optimization} shows the evolution of the continuous loss and 
the true AUC on the
training and test sets along the iteration for different values of $n_r$,
from $n_r=1$ (repartition at each iteration) to $n_r=+\infty$ (no repartition).
The lines are the median at each iteration over 100
runs, and the shaded area correspond to confidence intervals for
the AUC and loss value of the testing dataset.
We can clearly see the benefits of repartition: without it, the median
performance is significantly lower and the variance across runs is very large.
The results also show that occasional repartitions (e.g.,
every 25 iterations) are sufficient to mitigate these issues significantly.


\section{Future Work}\label{sec:conclusion}

We envision several further research questions on the topic of distributed
tuplewise learning. We would like to provide a rigorous convergence rate
analysis of the general distributed SGD algorithm introduced in 
\cref{sec:erm}. This is a challenging task because each series of
iterations executed between two repartition steps can be seen as optimizing
a slightly different objective function.
It would also be interesting to investigate settings where
the workers hold sensitive
data that they do not want to share in the clear due to privacy concerns.







%
%
%
\bibliographystyle{abbrv} 
\bibliography{spark-pairs,ERM_sampling}

\newpage
\appendix

\section*{SUPPLEMENTARY MATERIAL}


The code of the experiments can be found on the authors' repository.\footnote{
\url{https://github.com/RobinVogel/Trade-offs-in-Large-Scale-Distributed-Tuplewise-Estimation-and-Learning}}

\section{Acknowledgments}
This work was supported by IDEMIA. We would like to thank Anne Sabourin for
her feedback that helped improve this work, as well as the
ECML PKDD reviewers for their constructive input.

\section{Proof of \cref{prop-SWOR-naive-est}}
\label{prop-SWOR-naive-est-proof}
First, consider $\mathrm{Var}(U_{\mathbf{n},N})$. 
Hoeffding's decomposition implies that:
\begin{align*}
U_{\mathbf{n},N}(h) -  U(h) = T_n(h) + T_m(h)+ 
\frac{1}{N} \sum_{k=1}^N \frac{1}{n_0 m_0} \sum_{i \in \mathcal{R}_k^\X} 
\sum_{j \in \mathcal{R}_k^\Z} h_0(X_i,Z_j) ,
\end{align*}
as well as the following properties,
$\forall k,l \in \{1, \dots, n\} \times \{ 1, \dots, m\}$,
\begin{align}
    \begin{split}
       \text{Cov} ( h_1(X_k), h_2(Z_l)) &= 0,\\ 
       \text{Cov} ( h_1(X_j), h_0(X_k,Z_l)) &= 0, \quad \forall j \in \{ 1, \dots, n \},\\
       \text{Cov} ( h_2(Z_j), h_0(X_k,Z_l)) &= 0, \quad \forall j \in \{ 1, \dots, m \},
    \end{split}\label{covar_var_results}
\end{align}
which imply the result. The variance of the complete U-statistic $U_{\mathbf{n}}$ is just the 
special case $N=1$ of the variance $U_{\mathbf{n},N}$. Explicitely,
\begin{align*}
    \text{Var}(U_{n,N}(h)) = \frac{\sigma_1^2}{n} + \frac{\sigma_2^2}{m}
    + \frac{N\sigma_0^2}{nm}.
\end{align*}
Now for $\widetilde{U}_{\mathbf{n},N,B}(h)$, since $\widetilde{U}_{\mathbf{n},N,B}$ conditioned upon
the data has expectation $U_{\mathbf{n},N}(h)$, i.e.
\begin{align*}
\E \left [ \widetilde{U}_{\mathbf{n},N,B}(h) \vert \mathcal{D}_n , \mathcal{Q}_m, (\mathcal{R}_k)_{k=1}^N \right ] &= U_{\mathbf{n},N}(h), 
\end{align*}
the law of total variance implies,
\begin{align*}
    \text{Var} (\widetilde{U}_{\mathbf{n},N,B}(h)) & = 
    \text{Var}( U_{\mathbf{n},N}(h) ) 
    + \E[\text{Var}( \widetilde{U}_{\mathbf{n},N,B}(h) 
    \vert \mathcal{D}_n , \mathcal{Q}_m, (\mathcal{R}_k)_{k=1}^N )],\\
    &= \text{Var}( U_{\mathbf{n},N}(h) )
    + \frac{1}{N}\E[\text{Var}( \widetilde{U}_{\mathcal{R}_1,B}(h) 
    \vert \mathcal{D}_n , \mathcal{Q}_m, (\mathcal{R}_k)_{k=1}^N )],\\
    & \text{(Since the draws of } B \text{ pairs on different workers are independent)}\\
    &= \text{Var}( U_{\mathbf{n},N}(h) ) + 
    \frac{1}{N} \left [ - \frac{1}{B} \text{Var}(U_{\mathcal{R}_1}) 
    + \frac{1}{B} \text{Var}(h(X,Z)) \right ],\\
    & \text{(See \cite{CBC2016})}\\
    &= \left( 1 - \frac{1}{B} \right ) \text{Var}( U_{\mathbf{n},N}(h) )
    + \frac{1}{NB} \text{Var}(h(X,Z)),
\end{align*}
which concludes our proof. Explicitly,
\begin{align*}
    \text{Var}(\widetilde{U}_{\mathbf{n},N,B}(h)) &=
    \left( 1 - \frac{1}{B} \right) 
    \left( \frac{\sigma_1^2}{n} + \frac{\sigma_2^2}{m} + \frac{N\sigma_0^2}{nm} \right)
    + \frac{1}{NB} \text{Var}\left( h(X,Z) \right).
\end{align*}

\section{Proof of \cref{prop-SWOR-redist-est}}
\label{prop-SWOR-redist-est-proof}

We first detail the derivation of $\text{Var}(\widehat{U}_{n,N,T}(h))$.  Define the
Bernouilli r.v. $\epsilon_i^t(k)$ as equal to one if $X_k$ is in partition $i$ at
time $t$, and similarly $\gamma_i^t(l)$ is equal to one if $Z_l$ is in
partition
$i$ at time $t$.  Note that for $t \ne t_1$, $\epsilon_i^t(k)$ and
$\epsilon_{i_1}^{t_1}(k_1)$ are independent, as well as  $\gamma_i^t(l)$ and
$\gamma_{i_1}^{t_1}(l_1)$.  Additionally,  $\epsilon_i^t(k)$ and
$\gamma_{i_1}^{t_1}(l)$ are independent for any $t,t_0 \in \{ 1,\dots, T\}^2$.

Hoeffding's decomposition implies:
\begin{align*}
    U_{\textbf{n}, N}^t(h) - U_n(h) &= \frac{1}{N} \sum_{i=1}^N \frac{1}{nm} \sum_{k=1}^n \sum_{l=1}^m (N^2 \epsilon_i^t(k) \gamma_i^t(l) - 1) h_0(X_k, Z_l).
\end{align*}
The law of total variance, the fact that conditioned upon the data $\widehat{U}_{\textbf{n}, N, T}(h)$ is an average of $T$ independent experiments and 
the properties of \cref{covar_var_results} imply:
\begin{align}
    \text{Var}\left( \widehat{U}_{\textbf{n}, N, T}(h) \right) 
    &= \text{Var}\left(U_n(h)\right)  + \E\left[ \text{Var} \left( \widehat{U}_{\textbf{n}, N, T}(h)  \vert \mathcal{D}_n , \mathcal{Q}_m \right) \right], \nonumber \\
    &= \text{Var}\left(U_n(h)\right) 
    + \frac{1}{T}\E\left[ \text{Var} \left( U_{\textbf{n}, N}^t(h)  \vert \mathcal{D}_n , \mathcal{Q}_m \right) \right], \nonumber \\
    &= \text{Var}\left(U_n(h)\right) + \frac{N^2 \sigma_0^2}{nmT} \sum_{i_1,i_2=1}^N \text{Cov} \left( \epsilon_{i_1}^t (1) \gamma_{i_1}^t (1), \epsilon_{i_2}^t (1) \gamma_{i_2}^t(1) \right).
    \label{eqn_0}
\end{align}
On the other hand, observe that:
\begin{align}
    \text{Cov} \left( \epsilon_{i_1}^t (1) \gamma_{i_1}^t (1), \epsilon_{i_2}^t (1) \gamma_{i_2}^t(1) \right)
    =\begin{cases}
        - N^{-4} \quad & \text{if } i_1 \ne i_2, \\
        N^{-2} - N^{-4} \quad & \text{if } i_1 = i_2.
    \end{cases}
    \label{eqn_1}
\end{align}
The result is obtained by plugging \cref{eqn_1} in \cref{eqn_0}. Explicitly,
\begin{align*}
    \text{Var}\left( \widehat{U}_{\textbf{n}, N, T}(h) \right) 
    &= \text{Var}\left(U_n(h)\right) + \frac{N-1}{nmT} \sigma_0^2.
\end{align*}

Using that $\E [ \widetilde{U}_{\mathbf{n},N,B,T} (h) \vert \mathcal{D}_n, \mathcal{Q}_m, \epsilon, \gamma] = \widehat{U}_{\mathbf{n},N,T}(h)$, 
we now compute $\text{Var}( \widetilde{U}_{\mathbf{n},N,B,T}(h) )$ by decomposing it as the variance
of its conditional expectation plus the expectation of its conditional variance. It writes:
\begin{align*}
\text{Var} ( \widetilde{U}_{\mathbf{n},N,B,T}(h) )
&=  \text{Var}\left(\widehat{U}_{\mathbf{n},N,T}(h) \right) 
+ \E\left[ \text{Var}\left( \widetilde{U}_{\mathbf{n},N,B,T} \vert \mathcal{D}_n, \mathcal{Q}_m, \epsilon, \gamma \right) \right] \\
&=  \text{Var}\left(\widehat{U}_{\mathbf{n},N,T}(h) \right) 
+ \frac{1}{NT} \E\left[ \text{Var}\left( \widetilde{U}_{B, \mathcal{R}_i^t} \vert \mathcal{D}_n, \mathcal{Q}_m, \epsilon, \gamma \right) \right] \\
& \text{(Since the draws of } B \text{ pairs on different workers are independent)}\\
&=  \text{Var}\left(\widehat{U}_{\mathbf{n},N,T}(h) \right) 
+ \frac{1}{NT} \left[ - \frac{1}{B} \text{Var} \left ( U_{\mathcal{R}_i^t} \right ) + \frac{1}{B} \text{Var}(h(X,Z)) \right] \\
& \text{(See \cite{CBC2016}.)} \\
&=  \frac{\text{Var} (h(X,Z))}{NTB}  + \left( 1 - \frac{1}{TB} \right) \left( \frac{\sigma_1^2}{n} + \frac{\sigma_2^2}{m} \right)
 + \frac{\sigma_0^2}{nm} \left [ 1 + \frac{N-1}{T} - \frac{N}{TB} \right],  
 \\
\end{align*}
which gives the desired result after reorganizing the terms.

\section{Why $\widehat{U}_{\mathbf{n},N,T}$ Dominates
$\widetilde{U}_{\mathbf{n},N,B,T}$ for prop-SWOR}\label{same_no_pairs_comp}

To establish a fair comparison between both estimators, we calculate the difference $\Delta$
between the variance of $\widehat{U}_{\mathbf{n},N,T_0}$ and
$\widetilde{U}_{\mathbf{n},N,B,T}$ for the same number of pairs, i.e. when $NBT = T_0 nm/N$. 
Note that $\widetilde{U}_{\mathbf{n},N,B,T}$ involves more 
repartitioning of the data in all sensible cases, i.e. as soon as $B < nm/N^2$.

The expressions of \cref{prop-SWOR-naive-est} and \cref{prop-SWOR-redist-est} imply:
\begin{align*}
    \Delta :=& \text{Var}\left( \widetilde{U}_{\mathbf{n}, N, B, T} (h) \right) 
    - \text{Var} \left( \widehat{U}_{n,N,T_0} \right),\\
    =& \sigma_0^2 \left[ \frac{N-1}{nm} \left( \frac{1}{T} - \frac{1}{T_0} \right) 
    - \frac{N}{nmTB} + \frac{1}{NTB}\right] 
    + \frac{\sigma_1^2}{TB} \left[ \frac{1}{N} - \frac{1}{n} \right] 
    + \frac{\sigma_2^2}{TB} \left[ \frac{1}{N} - \frac{1}{m} \right].
\end{align*}
Pluging in the constraint on the pairs gives:
\begin{align*}
    \Delta =& \sigma_0^2 \left[ \frac{N-1}{nmT} \left( 1-\frac{1}{B} \right)
    + \frac{1}{TB} \left( \frac{1}{N^2} - \frac{1}{nm} \right)\right] 
    + \frac{\sigma_1^2}{TB} \left[ \frac{1}{N} - \frac{1}{n} \right] 
    + \frac{\sigma_2^2}{TB} \left[ \frac{1}{N} - \frac{1}{m} \right],
\end{align*}
which implies that $\Delta > 0$.

\section{Empirical Results for Sampling Without Replacement (SWOR)}
\label{SWOR_emp}

\begin{figure}[t]
 \centering
\begin{minipage}[t]{0.4\linewidth}
 \includegraphics[width=1.0\linewidth]{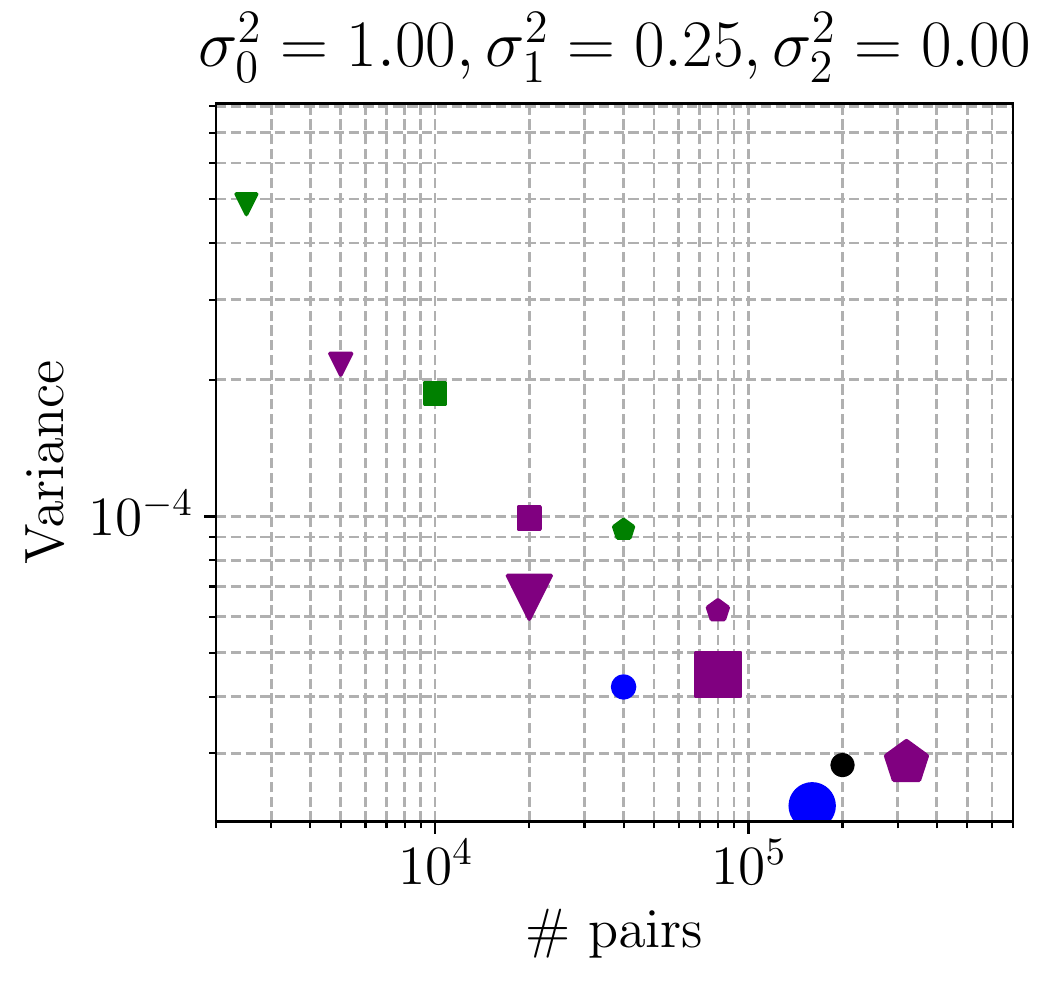}
\end{minipage}%
\hspace{.5cm}
\begin{minipage}[t]{0.4\linewidth}
 \includegraphics[width=1.0\linewidth]{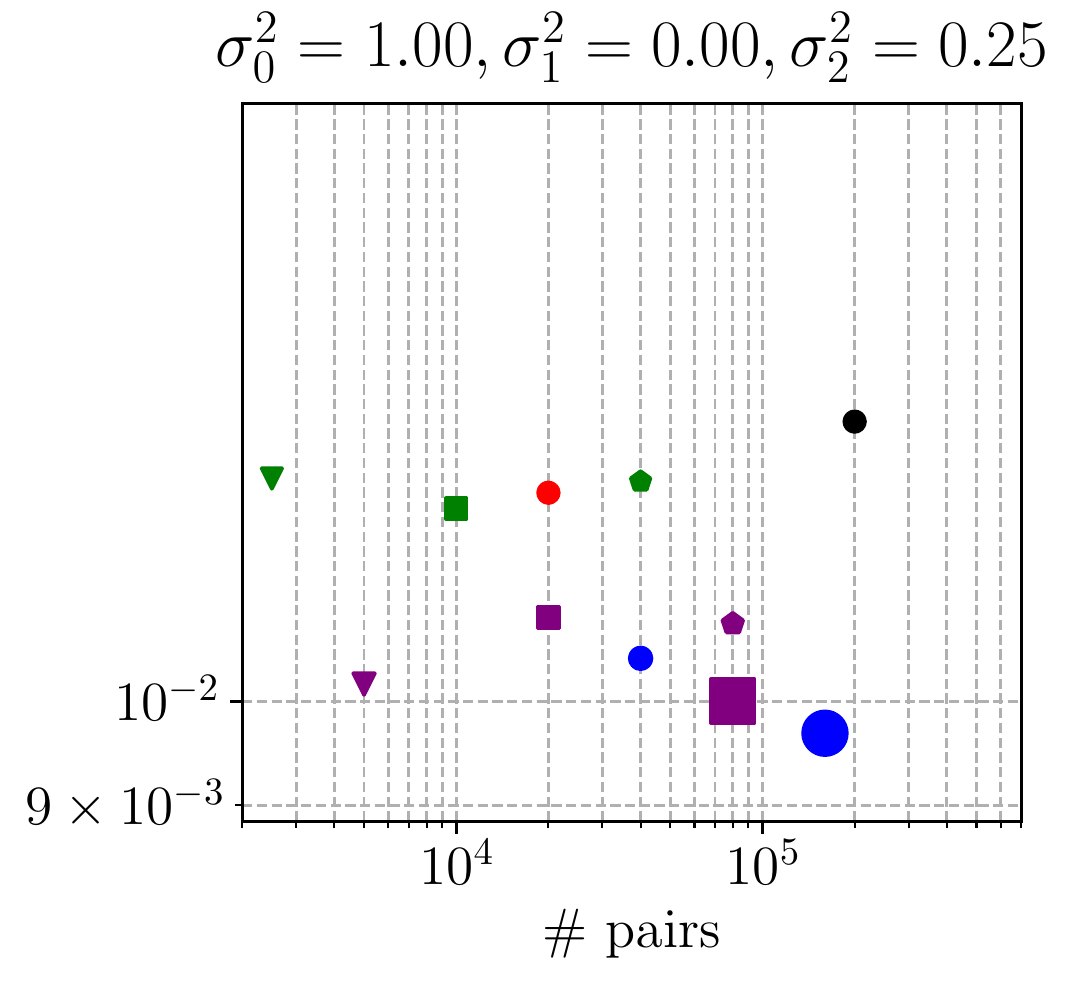}
\end{minipage}
\includegraphics[width=0.7\linewidth]{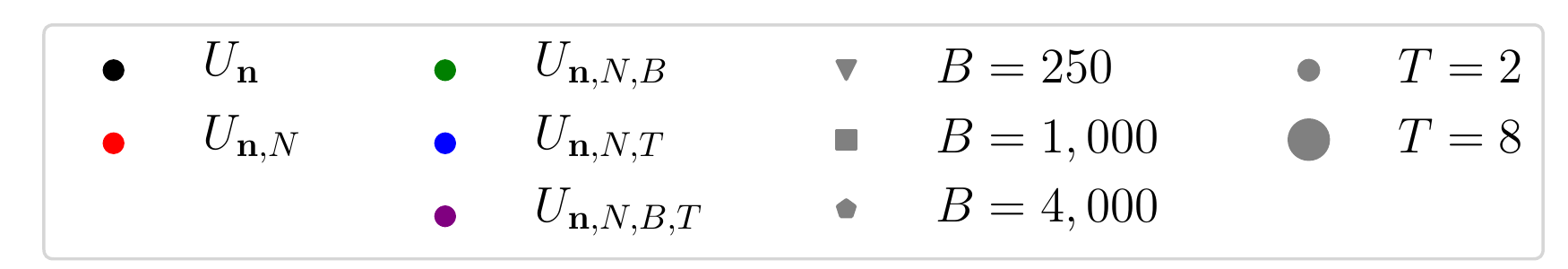}
\caption{Empirical variances as a function of the number of evaluated pairs
    for SWOR, with $n=100,000$, $m=200$ and $N=100$, evaluated over
    500 runs.}\label{fig:SWOR_emp_fig}
\end{figure}

In this section, we numerically show that in practice, the results for SWOR do
not deviate much from the theoretical ones obtained for prop-SWOR in 
\cref{prop-SWOR-naive-est} and \cref{prop-SWOR-redist-est}.
To illustrate this, we use the kernel $h(x,z) = x\cdot z$ and random variables
in $\R$ that follow a normal law $X \sim \mathcal{N} (\mu_X, \sigma_X)$ and
$Z \sim \mathcal{N}(\mu_Z, \sigma_Z)$. In that setting, note that 
$\sigma_1^2 = \mu_Z^2 \sigma_X^2$, $\sigma_2^2 = \mu_X^2 \sigma_Z^2$ and
$\sigma_0^2 = \sigma_X^2 \sigma_Z^2$, which means that by tweaking the
parameters $\mu_X, \mu_Z, \sigma_X, \sigma_Z$, one can obtain any possible value of
$\sigma_1, \sigma_2, \sigma_0$.

The results, shown in \cref{fig:SWOR_emp_fig} are
very similar to those obtained for prop-SWOR in \cref{fig:SWOR_theo}. The
fact that SWOR has slightly lower variance is expected,
since when no pairs are available the default value is always 0. This makes
the estimator give a stable prediction, but also makes it biased.

\section{Analysis of Proportional Sampling with Replacement (prop-SWR)}
\label{SWR_analysis}

While the use of prop-SWR is not very natural in a standard distributed
setting, it is relevant in cases where workers have access to joint database
that they can efficiently subsample.
We have the following results for the variance of estimates based on prop-SWR 
(see \cref{proof-prop-SWR-naive-est}
and \cref{proof-prop-SWR-redist-est} for the proofs).

\begin{theorem}
    If the data is distributed between workers with prop-SWR, and denoting $
    \mathbf{n}_0 = (n/N, m/N)$, we have:
\begin{align*}
    \text{Var}(U_{\mathbf{n}, 1}(h)) &= \frac{\sigma_1^2}{n} \left( 2 - \frac{1}{n} \right ) 
    + \frac{\sigma_2^2}{m} \left( 2 - \frac{1}{m} \right)
    + \frac{\sigma_0^2}{nm} \left[ 4 - 2 \left( \frac{1}{n} 
    + \frac{1}{m} \right) + \frac{1}{nm}\right], \\
    \text{Var}(U_{\mathbf{n}, N}(h)) &= \text{Var}(U_{\mathbf{n}, 1}(h))
    + \frac{\sigma_0^2}{nm} \left( N-1 \right)\left( 1 - \frac{1}{n} \right) \left( 1 - \frac{1}{m} \right), \\
    \text{Var}(\widetilde{U}_{\mathbf{n},N,B}(h)) &=  \text{Var} \left( U_{\textbf{n}, N}(h) \right) 
     + \frac{1}{NB} \left[ \sigma^2 - \text{Var} \left( U_{\mathbf{n}_0, 1}(h) \right) \right].
\end{align*}
\label{prop-SWR-naive-est}
\end{theorem}

\begin{theorem}
If the data is distributed and repartitioned between workers with prop-SWR, we
have:
\begin{align*}
    \text{Var}(\widehat{U}_{\mathbf{n},N,T}(h)) &= \text{Var}(U_{\mathbf{n}}(h)) 
    + \frac{1}{T} \left[ \text{Var}(U_{\mathbf{n}, N}(h)) - \text{Var}(U_{\mathbf{n}}(h))  \right],\\
    \text{Var}(\widetilde{U}_{\mathbf{n},N,B,T}(h)) &= \text{Var}\left( \widehat{U}_{\mathbf{n},N,T}(h) \right) 
    + \frac{1}{NBT} \left[ \sigma^2 - \text{Var}(U_{\mathbf{n}_0, 1}(h))\right ].
\end{align*}
\label{prop-SWR-redist-est}
\end{theorem}

\begin{figure}[t]
 \centering
\begin{minipage}[t]{0.49\linewidth}
 \includegraphics[width=1.0\linewidth]{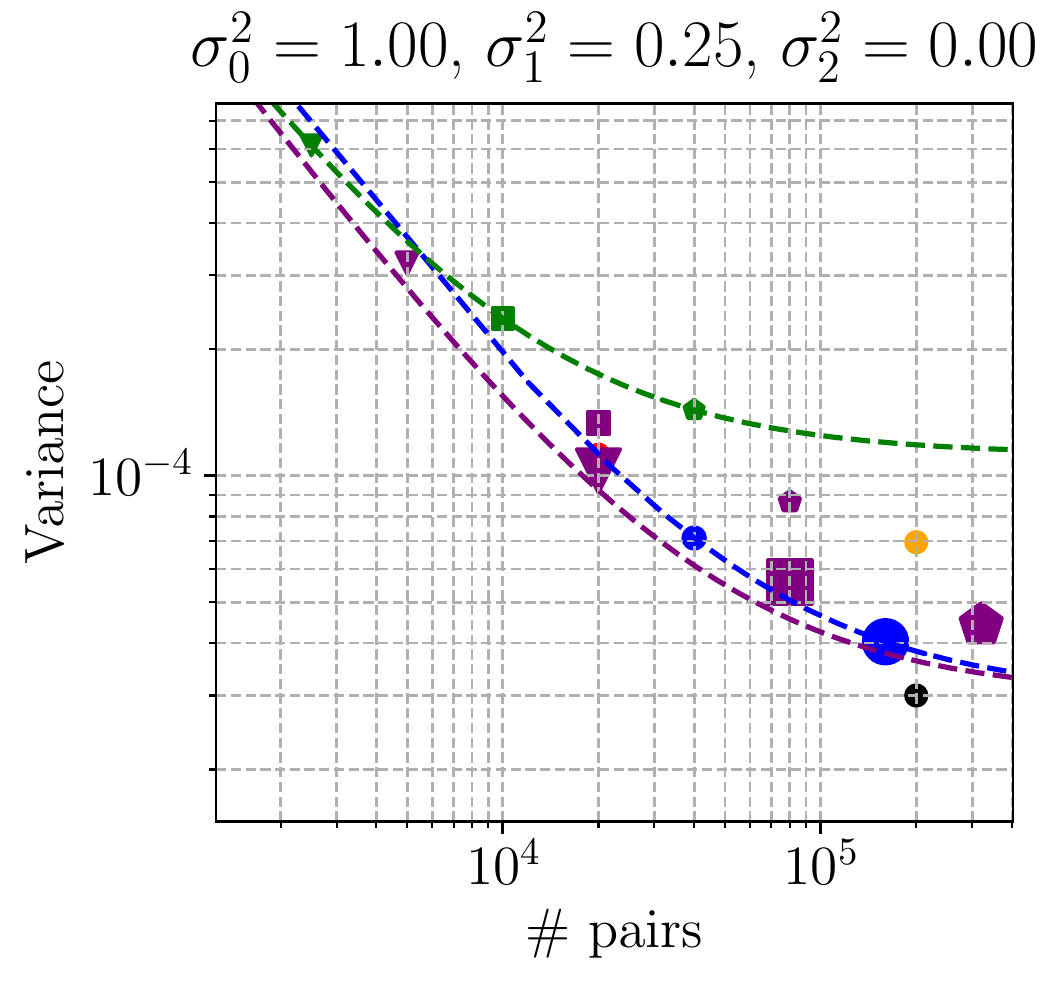}
\end{minipage}%
\hfill
\begin{minipage}[t]{0.49\linewidth}
 \includegraphics[width=1.0\linewidth]{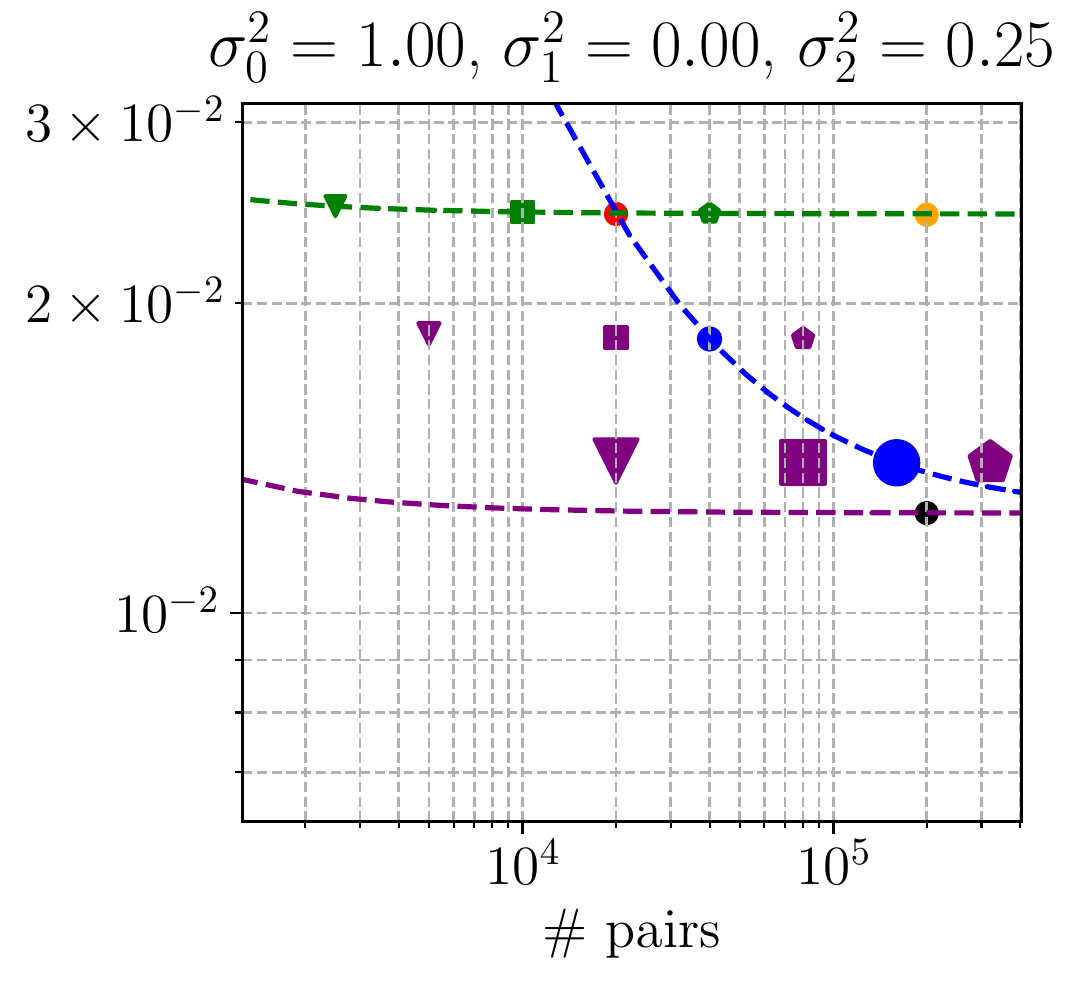}
\end{minipage}
 \includegraphics[width=1.0\linewidth]{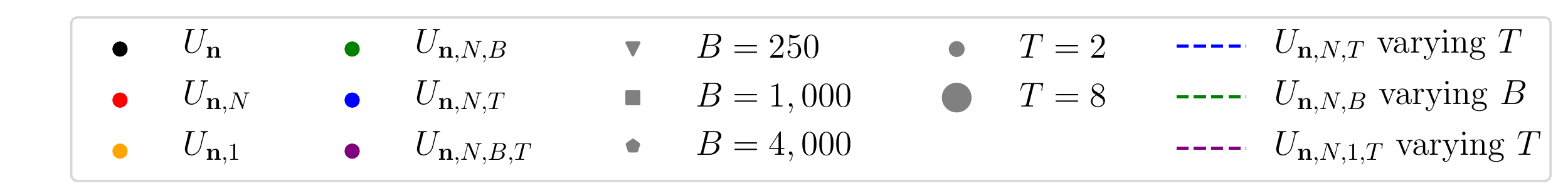}
\caption{Theoretical variances as a function of the number of evaluated pairs
    for different estimators under prop-SWR, with $n=100,000$, $m=200$
    and $N=100$.}\label{fig:SWR_theo}
\end{figure}

\cref{fig:SWR_theo} gives a visual illustration of these results. First note
that they are similar to those obtained for prop-SWOR in 
\cref{fig:SWOR_theo}. Yet, the right-hand
side figure shows that $\widetilde{U}_{\mathbf{n}, N, B, T}$ can have a
significantly lower variance than $\widehat{U}_{\mathbf{n}, N, T}$, for the
same number of evaluated pairs. This comes from the fact that
$\widetilde{U}_{\mathbf{n}, N, B, T}$ works on more bootstrap
re-samples of the data than $\widehat{U}_{\mathbf{n}, N, T}$, hence better
correcting for the loss of information due to sampling with replacement (at
the cost of more communication or disk reads). To stress this, we also
represented $U_{
\mathbf{n},1}$, i.e.  the point that gives the variance of a complete
estimator based on one bootstrap re-sample of the data.

\subsection{Proof of \cref{prop-SWR-naive-est}}
\label{proof-prop-SWR-naive-est}
First we derive the variance of $U_{\mathbf{n},N}(h)$.
Since $\E [U_{\mathbf{n},N}(h) \vert \mathcal{D}_n, \mathcal{Q}_m] = U_{\mathbf{n}}(h)$, the law of total variance implies:
\begin{align*}
\text{Var} ( U_{\mathbf{n},N}(h)) &= \text{Var}(U_{\mathbf{n}}(h)) + \E \left [ \text{Var}\left( U_{\mathbf{n},N}(h) \vert \mathcal{D}_n, \mathcal{Q}_m \right)\right ],\\
&= \text{Var}(U_{\mathbf{n}}(h)) + \frac{1}{N} \E \left [ \text{Var}\left( U_{\mathcal{R}_1}(h) \vert \mathcal{D}_n, \mathcal{Q}_m \right)\right ].
\end{align*}
Introduce $\epsilon(k)$ (resp. $\gamma(l)$) as the random variable that is equal to
the number of times that $k$ has been sampled in cluster $1$ for the
$\mathcal{D}_n$ elements (resp. that $l$ has been sampled in cluster $1$ for
the $\mathcal{Q}_m$ elements).  The random variable $\epsilon(k)$ (resp.
$\gamma(l)$) follows a binomial distribution with parameters $(n/N,1/n)$
(resp. $(m/N, 1/m)$). Note that the $\epsilon$ and $\gamma$ are independent
and that $\sum_{k=1}^n \epsilon(k) = n/N$ and $\sum_{l=1}^m \gamma(l) = m/N$.  
It follows that:
\begin{align*}
    U_{\mathcal{R}_1}(h) - U_n(h)=   U(h)
    & + \frac{1}{n}\sum_{k=1}^n \left( N\epsilon(k)-1 \right) \left( h_1(X_k) - U(h) \right)  \\
    & + \frac{1}{m}\sum_{l=1}^m \left( N\gamma(l) - 1 \right) \left( h_2(Z_l) - U(h) \right)  \\
    & + \frac{1}{nm} \sum_{k=1}^n \sum_{l=1}^m \left( N^2 \epsilon(k) \gamma(l) - 1 \right)h_0(X_k, Z_l),
\end{align*}
which implies, using the results of \cref{covar_var_results},
\begin{align}
    \E\left[\text{Var}\left( U_{\mathcal{R}_1}(h) \vert \mathcal{D}_n, \mathcal{Q}_m \right) \right] &= 
    \frac{N^2\sigma_1^2}{n} \text{Var}(\epsilon(1)) + \frac{N^2\sigma_2^2}{m} \text{Var}(\gamma(1)) +
    \frac{N^4\sigma_0^2}{nm} \text{Var}(\epsilon(1) \gamma(1)).
    \label{inter_var}
\end{align}
The mean and variance of a binomial distribution is known. Since $\epsilon(1)$ and $\gamma(1)$ are independent,
\begin{align}
\begin{split}
    &\text{Var}(\epsilon(1) \gamma(1)) = \frac{1}{N^2}\left[ \left( 1-\frac{1}{n} \right) \left( 1-\frac{1}{m} \right) 
    + \frac{1}{N} \left( 2 - \frac{1}{n} - \frac{1}{m} \right) \right], \\
    &\text{Var}(\epsilon(1)) = \frac{1}{N} \left( 1-\frac{1}{n} \right), 
    \quad \text{Var}(\gamma(1)) = \frac{1}{N} \left( 1-\frac{1}{m} \right).
\end{split}\label{variances_expressions}
\end{align}
Plugging \cref{variances_expressions} into \cref{inter_var} gives the result. Explicitly,
\begin{align*}
    \text{Var} \left( U_{\textbf{n}, N}(h) \right) =&
    \frac{\sigma_1^2}{n} \left( 2-\frac{1}{n} \right)
    + \frac{\sigma_2^2}{m} \left( 2-\frac{1}{m} \right) \\
    & + \frac{\sigma_0^2}{nm} \left[ \left( 3 - \frac{1}{n} - \frac{1}{m} \right) + N \left( 1 - \frac{1}{n} \right) \left( 1 - \frac{1}{m} \right)  \right].
\end{align*}

Now we derive the variance of $\widetilde{U}_{\mathbf{n},N,B}(h)$.
Note that $\E[\widetilde{U}_{\mathbf{n},N,B} \vert \mathcal{D}_n, \mathcal{Q}_m , \epsilon , \gamma ] = U_{\mathbf{n},N}(h)$, hence:
\begin{align}
    \text{Var}&(\widetilde{U}_{\mathbf{n},N,B}) = \text{Var}(U_{\mathbf{n},N}(h)) + 
    \E \left[ \text{Var} \left( \widetilde{U}_{\mathbf{n},N,B} 
    \vert \mathcal{D}_n, \mathcal{Q}_m , \epsilon , \gamma \right ) \right], \nonumber\\
    &= \text{Var}(U_{\mathbf{n},N}(h)) + \frac{1}{N} 
    \E \left[ \text{Var} \left( \widetilde{U}_{B,\mathcal{R}_1} \vert 
    \mathcal{D}_n, \mathcal{Q}_m , \epsilon , \gamma \right ) \right].
    \label{eq_11}
\end{align}
Conditioned upon $\mathcal{D}_n, \mathcal{Q}_m , \epsilon , \gamma$, the statistic $\widetilde{U}_{B,\mathcal{R}_1}$
is an average of $B$ independent experiments. Introducing $\delta_{k,l}$ as equal to $1$ if the pair $(k,l)$ is 
selected in worker $1$ as the $1$th pair of $\widetilde{U}_{B,\mathcal{R}_1}$, 
and $\Delta_{k,l}$ its expected value, i.e. $\Delta_{k,l} := \E[\delta_{k,l}] = N^2 \epsilon(k) \gamma(l)/nm$, it implies
\begin{align}
 \text{Var} &\left( \widetilde{U}_{B,\mathcal{R}_1} \mid \mathcal{D}_n, \mathcal{Q}_m , \epsilon , \gamma \right )
 = \frac{1}{B} \text{Var}\left(\sum_{k=1}^{n} \sum_{l=1}^m  \delta_{k,l} h(X_k,Z_l) \mid \mathcal{D}_n, \mathcal{Q}_m , \epsilon , \gamma \right).
 \label{eq_12}
\end{align}
From the definition of $\delta_{k,l}$ we have $\delta_{k,l} \delta_{k_1, l_1}=0$ as soon as $k \ne k_1$ or $l \ne l_1$, writing the 
right-hand-side of \cref{eq_12} as the second order moment minus the squared means gives:
\begin{align}
 \text{Var} &\left( \widetilde{U}_{B,\mathcal{R}_1} \mid \mathcal{D}_n, \mathcal{Q}_m , \epsilon , \gamma \right )
 = \frac{1}{B} \sum_{k=1}^{m} \sum_{l=1}^n  \Delta_{k,j} h^2(X_k,Z_l) -  \frac{1}{B} \left( \sum_{k=1}^{n} \sum_{l=1}^m \Delta_{k,l}  h(X_k,Z_l) \right)^2.
  \label{eq_13}
\end{align}
Taking the expectation of \cref{eq_13} gives:
\begin{align}
  \E \Big [\text{Var}  \left( \widetilde{U}_{B,\mathcal{R}_1} \vert \mathcal{D}_n, \mathcal{Q}_m , \epsilon , \gamma \right ) \Big ]  
  & = \frac{1}{B} \left[ \E[h^2(X,Z)] - \E[U_{\mathcal{R}_1}^2] \right], \nonumber\\
  & = \frac{1}{B} \left[ \text{Var}(h(X,Z)) - \text{Var}(U_{\mathcal{R}_1}) \right ].
  \label{eq_14}
\end{align}
Pluging \cref{eq_14} into \cref{eq_11} gives
\begin{align*}
\text{Var}(\widetilde{U}_{\mathbf{n},N,B}) &=
\frac{\text{Var}(h(X,Z))}{BN}  + \text{Var}(U_{\mathbf{n},N}(h)) - \frac{\text{Var}(U_{\mathcal{R}_i})}{BN} ,
\end{align*}
and we can conclude from preceding results, since $U_{\mathcal{R}_i}$ is simply $U_{\mathbf{n}_0, 1}$ with $n_0 = (n/N, m/N)$.

\subsection{Proof of \cref{prop-SWR-redist-est}}
\label{proof-prop-SWR-redist-est}

Since $\E\left[ \widehat{U}_{\mathbf{n},N,T}(h) \vert \mathcal{D}_n, \mathcal{Q}_m\right] = U_{\mathbf{n}}(h)$
the law of total covariances followed by the fact that, conditioned upon $\mathcal{D}_n, \mathcal{Q}$, 
the statistic $\widehat{U}_{\mathbf{n},N,T}(h)$ is an average of $T$ independent random variables, implies:
\begin{align*}
\text{Var} \left(\widehat{U}_{\mathbf{n},N,T}(h)\right) = \text{Var}(U_{\mathbf{n}}(h)) 
+ \frac{1}{T} \E\left[ \text{Var} \left(U_{\mathbf{n},N}(h) \vert \mathcal{D}_n, \mathcal{Q}_m \right) \right].
\end{align*}
The calculations of \cref{proof-prop-SWR-redist-est} give the result. Explicitly,
\begin{align*}
    \text{Var} \left( \widehat{U}_{\mathbf{n},N,T}(h) \right) &=
    \text{Var}(U_{\mathbf{n}}(h)) + \frac{1}{T} \left[ \text{Var}(U_{\mathbf{n}, N}(h)) - \text{Var}(U_{\mathbf{n}}(h))  \right].
\end{align*}

We now derive the variance of $\widetilde{U}_{\mathbf{n},N,B,T}$.
Since 
\begin{align*}
    \E\left[ \widetilde{U}_{\mathbf{n},N,B,T}(h) \vert \mathcal{D}_n, \mathcal{Q}_m, \epsilon, \gamma \right] = \widehat{U}_{\mathbf{n},N,T}(h) ,
\end{align*}
the law of total covariance followed by the calculations of \cref{proof-prop-SWR-redist-est} imply the result:
\begin{align*}
\text{Var}\left(\widetilde{U}_{\mathbf{n},N,B,T} (h)\right) 
&= \text{Var}\left( \widehat{U}_{\mathbf{n},N,T}(h) \right) 
+ \E\left[ \text{Var} \left( \widetilde{U}_{\mathbf{n},N,B,T}(h) \vert \mathcal{D}_n, \mathcal{Q}_m, \epsilon, \gamma \right) \right] ,\\
&= \text{Var}\left( \widehat{U}_{\mathbf{n},N,T}(h) \right) 
+ \frac{1}{T}\E\left[ \text{Var} \left( \widetilde{U}_{\mathbf{n},N,B}(h) \vert \mathcal{D}_n, \mathcal{Q}_m, \epsilon, \gamma \right) \right] ,\\
&= \text{Var}\left( \widehat{U}_{\mathbf{n},N,T}(h) \right) 
+ \frac{1}{NBT} \left[ \text{Var}(h(X,Z)) - \text{Var}(U_{\mathcal{R}_i})\right ].
\end{align*}

\section{Details on the Estimation Experiment of \cref{sec:experiments}}

Here we give some details on the derivations leading to \cref{sigmas_ext_exp}.
We have:
\begin{align*}
    U(h) &= \Prob{X>Z} = q + (1-q)(1-p),\\
    h_1(x) &= \Prob{x>Z} = 1-p + p\cdot \I\{x=2\},\\
    h_2(z) &= \Prob{X>z} = q + (1-q) \I\{z=-1\},\\
    h_0(x,z) &= \I\{x=2\} + \I\{x=0\}\cdot \I\{z=-1\} - h_1(x) - h_2(z) + U(h) ,\\
    &= - \left( \I\left\{ X=2 \right\}-q \right) \left( \I\left\{ Z=-1 \right\} - (1-p) \right).
\end{align*}
It follows that:
\begin{align*}
    \sigma_1^2 & = p^2 q(1-q), \\
    \sigma_2^2 & = (1-q)^2 p(1-p), \\
    \sigma_0^2 & = pq (1-p)(1-q).
\end{align*}

\end{document}